\documentclass[10pt,journal,compsoc]{IEEEtran}

\usepackage[utf8]{inputenc}
\usepackage[english]{babel}
\usepackage{underscore}

\usepackage{amsfonts}
\usepackage{amsmath}
\usepackage{xcolor}

\usepackage[urlcolor=blue,linkcolor=red, citecolor =blue,colorlinks=true,bookmarks=false,hypertexnames=true]{hyperref}

\newcommand{\revise}[1]{{\color{black}{#1}}}  %

\newcommand{\ndrw}[1]{{\color{black}{#1}}}
\newcommand{\ndrww}[1]{{\color{black}{#1}}}  %
\newcommand{\denis}[1]{{\color{black}{#1}}}
\newcommand{\denisnew}[1]{{\color{black}{#1}}}  %
\newcommand{\dennis}[1]{{\color{black}{#1}}}

\newcommand{\maxim}[1]{{\color{black}{#1}}}

\ifCLASSOPTIONcompsoc
  \usepackage[nocompress]{cite}
\else
  \usepackage{cite}
\fi

\ifCLASSINFOpdf
  \usepackage[pdftex]{graphicx}
  \usepackage{subfig}
  \graphicspath{Figures}
\else
\fi

\addto\captionsenglish{%
  \renewcommand{\contentsname}%
    {Face Generation and Editing with StyleGAN: A Survey}%
}
\begin{document}

\onecolumn
\tableofcontents

\twocolumn

\title{Face Generation and Editing with StyleGAN: \\A Survey}

\author{
Andrew Melnik,
Maksim Miasayedzenkau,
Dzianis Makarovets,
Dzianis Pirshtuk,
Eren Akbulut,\\
Dennis Holzmann,
Tarek Renusch,
Gustav Reichert,
and~Helge Ritter%
    \IEEEcompsocitemizethanks{
    \IEEEcompsocthanksitem E-mail: andrew.melnik.papers@gmail.com
    \IEEEcompsocthanksitem
    A. Melnik, E. Akbulut, D. Holzmann, T. Renusch, G. Reichert, H. Ritter - Bielefeld Univeristy, Germany.
        
    \IEEEcompsocthanksitem M. Miasayedzenkau, D. Makarovets, D. Pirshtuk - Banuba
        }%
    }

\IEEEtitleabstractindextext{%
\begin{abstract}

Our goal with this survey is to provide an overview of the state of the art deep learning methods for face generation and editing using StyleGAN. The survey covers the evolution of StyleGAN, from PGGAN to StyleGAN3, and explores relevant topics such as suitable metrics for training, different latent representations, GAN inversion to latent spaces of StyleGAN, face image editing, cross-domain face stylization, face restoration, and even Deepfake applications. We aim to provide an entry point into the field for readers that have basic knowledge about the field of deep learning and are looking for an accessible introduction and overview.

\end{abstract}

}

\maketitle

\IEEEdisplaynontitleabstractindextext

\IEEEpeerreviewmaketitle

\IEEEraisesectionheading{\section{Introduction}\label{sec:introduction}}

\begin{figure}[!t]
    \centering
    \includegraphics[width=3.5in]{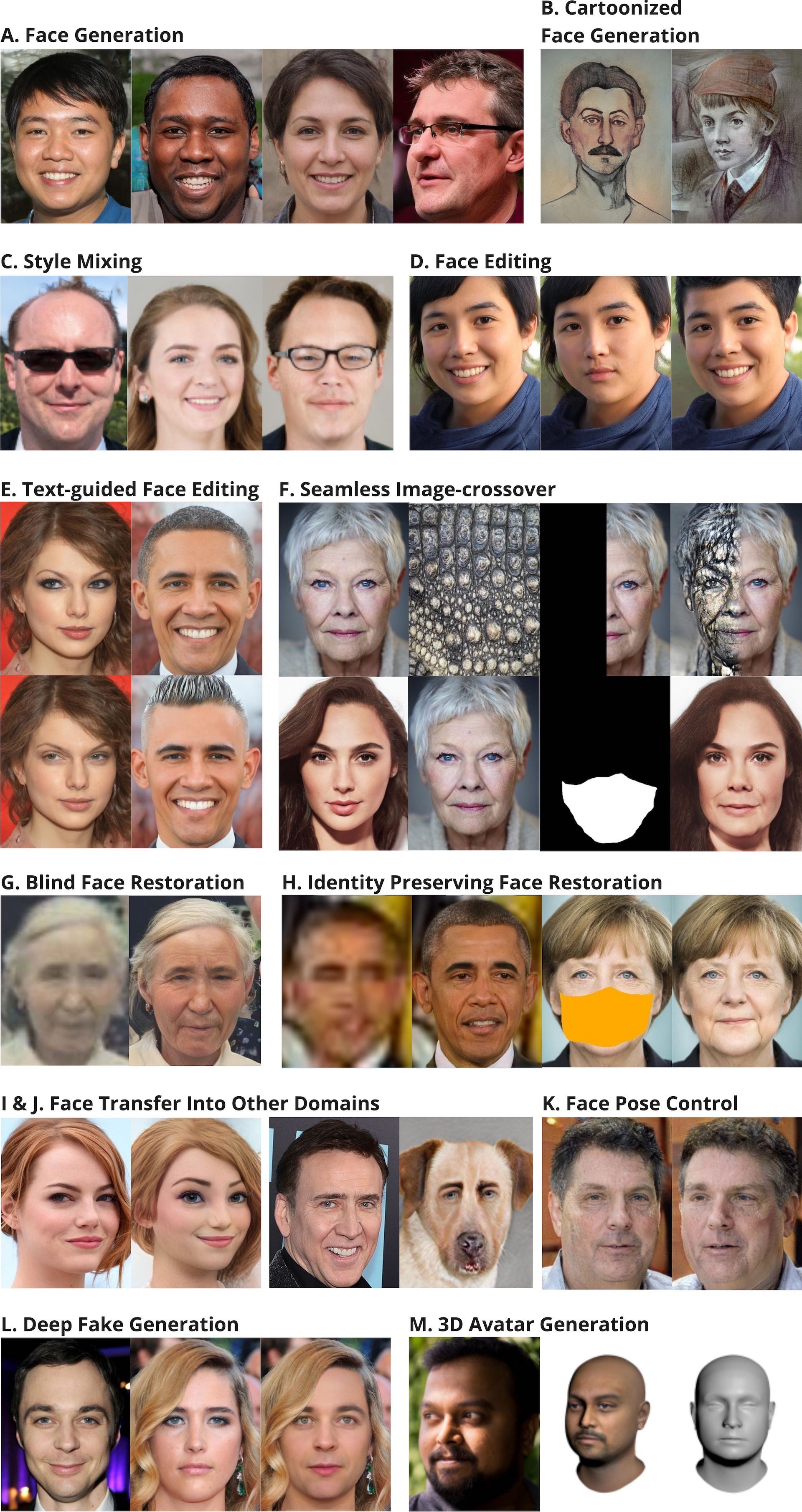}
    \vspace{-0.2in}
    \caption{
        \textbf{Synopsis of StyleGAN Applications}. \textbf{A}. Faces generated using StyleGAN2 \cite{karras2020analyzing}. \textbf{B}. NFT collection \cite{GANFolksAll} generated using StyleGAN2 \cite{karras2020analyzing}
        trained on MetFaces dataset \cite{karras2020training}. \textbf{C}. Style mixing (see Figure \ref{fig:style_mixing}). \textbf{D}. From left to right: the source image, smile removed, gender changed \cite{karras2020analyzing}. \textbf{E}. Image editing with StyleCLIP \cite{patashnik2021styleclip} using text prompts. Upper row: original images, lower row: edited images using text prompts -- Emma Stone (left), Mohawk Style (right). \textbf{F}. Seamless image-crossover \cite{abdal2020image2stylegan++}. Left: sources image, right: resulting image. \textbf{G}. Blind face restoration \cite{wang2021towards}. Left: degraded image, right: enhanced image. \textbf{H}. Identity preserving face restoration and in-painting \cite{nitzan2022mystyle} \cite{mystyleex}. \textbf{I} and \textbf{J}. Transferring faces into other domains while preserving the identity. Left: real image, right: cartoon like output \cite{Toonify} \cite{DBLP:journals/corr/abs-2108-00946} \textbf{K}. Control semantic parameters, such as face pose using StyleGAN component. Left: source image, right: pose changed \cite{tewari2020stylerig}. \textbf{L}. Deepfake generation \cite{Zhu2021OneSF}. \textbf{M}. Automated 3D avatar generation using StyleGAN component. From left to right: source image, avatar, face model \cite{luo2021normalized}.
    }
    \label{fig:image_grid_applications}
    \vspace{-0.3in}
\end{figure}

\IEEEPARstart{H}{umans} have always been fascinated with faces. It is how we recognize people, it is the main feature we watch out for when interacting with other people. 
This is reflected in the fact that we have a specialized region in our brain solely dedicated to the detection of face patterns and their subtle changes \cite{parvizi2012electrical}\cite{melnik2017systems}.

There are almost 8 billion people alive today and yet we can discern all of them by their face.
Attempting to delineate the great diversity of faces, \cite{FacialFeaturesReport} proposes 26 facial features as relevant for the description of faces (including many shape and size features, along with features for color and texture). Providing just three value levels for each of these features already results in $3^{26}>2.5 \cdot 10^{12}$ unique faces, which is approximately 300 times greater than the current global population. Thus, there is still a vast space for unique identification of individuals using facial features. 

Our goal with this survey is to provide an overview of the state of the art deep learning technologies for face generation and editing. We particularly focus on GAN-based architectures that have culminated in the StyleGAN approaches. These methods enable the generation of high-quality facial images and provide versatile tools for semantics editing while preserving the image's photographic quality. The StyleGAN architecture merits the attention of a broader readership because it provides an ecosystem for a variety of applications and is used as the basis for a large number of research works. For a condensed visual synopsis, see Figure \ref{fig:image_grid_applications} and Chapter \ref{ch:Applications}.

\subsection{Survey Overview}
The plan of the paper is as follows: Chapter \ref{ch:Applications} attempts to give an impression of the richness of applications of StyleGAN-based image processing methods. Chapter \ref{sec:NNarch4FaceGen} will explain Generative Adversarial Networks (GANs), delving specifically into the StyleGAN architectures for the generation of face images. Training such architectures needs suitable metrics that capture image similarity at different levels, which will be the topic of Chapter \ref{ch:sim}. Chapter \ref{chapter:latentspace} will discuss the different latent representations that form the basis of the controllable image editing. Chapter \ref{ch:Inversion} focuses on finding the latent representation of a given image. This prepares the ground for the methods reviewed in Chapter \ref{sec:editStyleGAN} to edit face images, and in Chapter \ref{ch:cross_domain} for cross domain face stylization. In Chapter \ref{ch:FaceRestoration} we look at some major approaches connected with face restoration and producing deepfakes in Chapter \ref{ch:deep_fake_chapter}. In Chapter \ref{ch:Alternative} we provide a concise overview of alternatives to StyleGAN for face generation and editing methods. The last Chapter \ref{ch:Conclusion} concludes with a short summary and outlook.

\subsection{Synopsis of StyleGAN Applications}\label{ch:Applications}

\textbf{Synthetic Face Generation:} The StyleGAN \cite{karras2019style}\cite{karras2020analyzing} architecture can generate faces which do not exist (see Figure \ref{fig:image_grid_applications}A). Applications of StyleGAN include the generation of unique pieces of art, including NFT collections
\cite{GANFolksAll}\cite{GANFolksMedium} (see Figure \ref{fig:image_grid_applications}B). \denisnew{There have been even some StyleGAN-based techniques developed to generate a child's facial image using parental face images as input \cite{li2023stylegene} \cite{lin2021styledna} \cite{qin2015tri}. \textit{Style mixing} technique \cite{styletransfer} allows creation of facial images that share features of several source images (see Figure \ref{fig:image_grid_applications}C) by combining StyleGAN internal representations of source images (see Figure \ref{fig:style_mixing} and Chapter \ref{ch:cross_domain}).
}

\textbf{Editing Facial Features:} 
While preserving a high image quality \cite{alaluf2022third}, the StyleGAN architecture \cite{karras2020analyzing}\cite{karras2021alias} allows to alter many features like age, hair, smile, etc. (see Figure \ref{fig:image_grid_applications}D and more details in Chapter \ref{sec:editStyleGAN} for details)\cite{melnik2022faces}.
The StyleGAN architecture can be used as a component for text-driven editing of images. StyleCLIP \cite{radford2021learning}\cite{patashnik2021styleclip} offers a manipulation of facial features using text prompts alone (see Figure \ref{fig:image_grid_applications}E and Chapter \ref{ch:StyleCLIP}).
The \textit{Image2StyleGAN++} framework \cite{abdal2020image2stylegan++} provides application examples of image editing using scribbles, inpainting, or crossover (see Figure \ref{fig:image_grid_applications}F).

\textbf{Facial Image Recovery:}
StyleGAN has been demonstrated to be able to restore face images with regard to degradation such as low resolution, noise, colorization of old photos, and even missing parts (see Chapter \ref{ch:FaceRestoration}). For example, GFP-GAN \cite{wang2021towards} or other works \cite{yang2021gan}\cite{Poirier-Ginter_2023_CVPR} uses a pretrained StyleGAN2 \cite{karras2020analyzing} as a component of the face restoration architecture (see Figure \ref{fig:image_grid_applications}G). However using such general facial priors for restoration of images can end up in identity loss of faces. MyStyle \cite{nitzan2022mystyle} tackles the problem of target identity, while recovering lost information (see Chapter \ref{ch:myStyle_Inpainting_SResolution} and Figure \ref{fig:image_grid_applications}H)

\denisnew{
\textbf{Stylization of Faces:}
Transferring face images in another domain such as sketches\cite{Zhang2020}\cite{8621606}\cite{8764602} or cartoonization\cite{jang2021stylecarigan}\cite{wang20223d}\cite{back2021fine} while preserving the person identity is a big challenge. The goal of stylization approaches  is to satisfy both identity preservation and perceptual features of a certain style (see Figure \ref{fig:image_grid_applications}I and \ref{fig:image_grid_applications}J for example applications and Chapter \ref{ch:cross_domain} for technical details).}

\denis{
\textbf{Deepfake generation} can be summed up as a face-swapping operation that is not being recognizable by human viewers (see Figure \ref{fig:image_grid_applications}L). Deepfakes are used by artists, social media platforms, in film, game, fashion, and entertainment industry \cite{nguyen2022deep}\cite{mirsky2021creation}\cite{MyHeritage}\cite{Verdoliva2020MediaFA}. 
Refer to Chapter \ref{ch:deep_fake_chapter} for a comprehensive exploration of how StyleGAN functionality can be harnessed for deepfake applications: face reenactment \cite{Yang2019UnconstrainedFE}\cite{9157131}, swapping \cite{Zhu2021OneSF}, and transfer \cite{moniz2018unsupervised}.

\textbf{3D Facial Avatar Generation}
by transferring a 2D human face image to a 3D avatar, while preserving identity \cite{luo2021normalized} (see Figure \ref{fig:image_grid_applications}M). Such 3D avatars can be useful e.g., in gaming, real time video filters, etc. 
}

\ndrww{
\label{sec:Mobile_Networks}
\textbf{Hands-on Applications and Mobile Networks:}
Face generation and editing technologies are popular for social networks, messengers, mobile and photo apps. In many cases, privacy issues motivate the localization of processing from cloud servers to mobile devices. Mobile architecture networks \cite{Howard2017MobileNetsEC} are constrained to being fast while consuming only little memory on the order of a few megabytes (MB) instead of several gigabytes. 
Running architectures that include the StyleGAN model can be computationally demanding. Thus, in many cases the StyleGAN architecture can be used to generate datasets of paired examples for supervised learning of a number of
mobile encoder-decoder networks (1-10 MB), one per each discrete editing feature like adding glasses, smile, makeup, etc \cite{li2020gan}\cite{viazovetskyi2020stylegan2}. 
}

\subsection{Training datasets}
\label{datasets}

The majority of the approaches discussed in this survey used the following datasets for training: Flickr-Faces-HQ (FFHQ) \cite{karras2019style}, CelebFaces Attributes Dataset (CelebA) \cite{liu2015deep}, and CelebA-HQ \cite{Karras2018ProgressiveGONA}.

\textbf{FFHQ} is a dataset of 70,000 high-quality PNG images at 1024×1024 resolution, featuring diverse human faces with variation in age, ethnicity, and accessories such as eyeglasses and hats. Originally designed for GAN benchmarking, the images were obtained from Flickr and were automatically aligned and cropped using dlib \cite{dlib09}, inheriting the website's biases.

\textbf{CelebA} is a face attributes dataset containing over 200,000 celebrity images at $178 \times 218$ resolution, each annotated with 40 attributes. The images have diverse backgrounds and poses and come with rich annotation, including 10,177 identities, 202,599 face images, 40 binary attribute annotations, and 5 landmark locations (eyes, nose, mouth) per image. \textbf{CelebA-HQ} is a high quality version of CelebA dataset with 30,000 images at $1024 \times 1024$ resolution.

\begin{figure*}
    \centering
    \includegraphics[width=\linewidth]{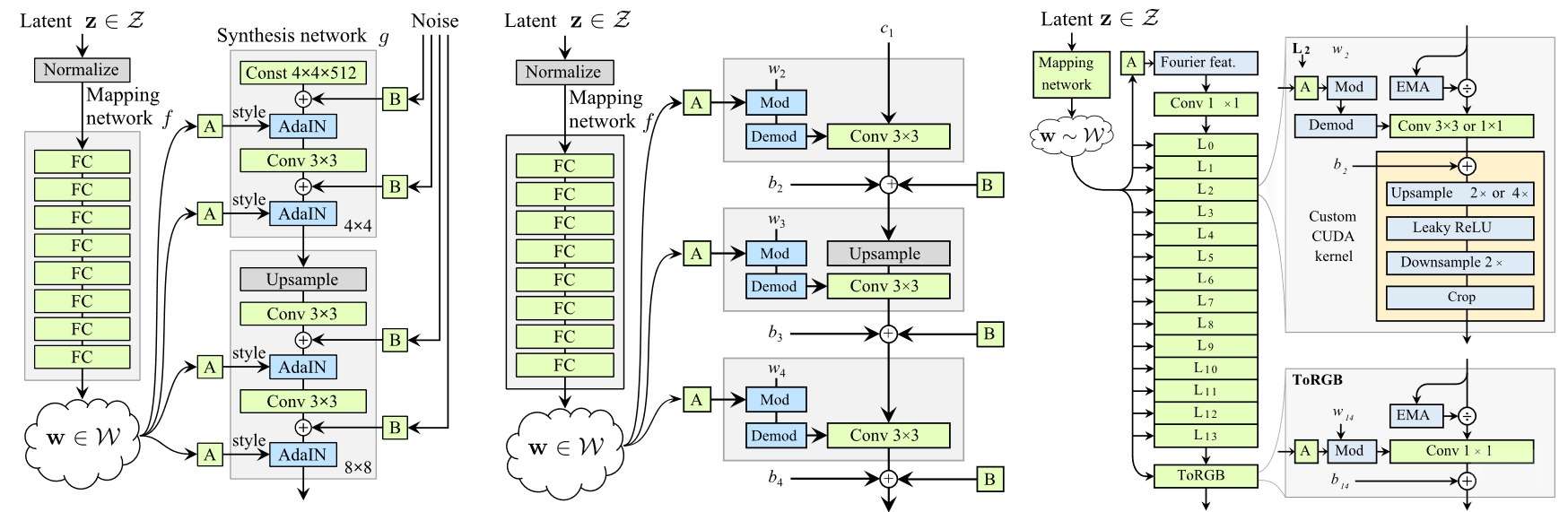}
    \caption{Architectures of StyleGAN generators: (a) StyleGAN \cite{karras2019style}, (b) StyleGAN2 \cite{karras2020analyzing}, (c) StyleGAN3 \cite{karras2021alias}}
    \vspace{-0.2in}
    \label{fig:gan_architectures}
\end{figure*}

\ndrw{

\section{StyleGAN Architectures for Generation of Faces}
\label{sec:NNarch4FaceGen}

\subsection{Generative Adversarial Networks}
\label{ch:gan}

To generate images of faces, Generative Adversarial Networks (GANs) \cite{goodfellow2020generative} have proven to be a highly suitable architecture. 
A GAN offers a way to learn a mapping that transforms a known, usually simple (e.g., Gaussian) distribution into a more complex target distribution that represents a given domain of patterns, for example, images of human faces. This mapping or generative model can be learned using a sufficiently large training set of samples from the desired target distribution. Once the model has been learned, new samples can be generated by simply feeding the model with random inputs from the simple distribution that was used during training.

\subsection{Progressive growing GANs (PGGAN)}

While vanilla GANs are able to generate images of reasonable quality, they suffer from limited controllability and unstable training \cite{convergenceGANs}. To overcome these problems, Karras et al. \cite{Karras2018ProgressiveGONA} introduced a training strategy in which the neural network progressively grows more layers during training (see Figure \ref{fig:gan_output}A). With layers at increasing depth, image resolution increases as well. Starting with low-resolution images of $4 \times 4$ up to a resolution of $1024 \times 1024$, firstly coarse structures and later fine details are learned. This makes training more stable, because it splits the task into simpler sub-tasks. Additionally, the training time benefits from this approach, because most of the iterations are done at lower resolutions and thus in a network with a smaller number of layers.

\begin{figure}[!t]
    \centering
    \includegraphics[width=3.0in]{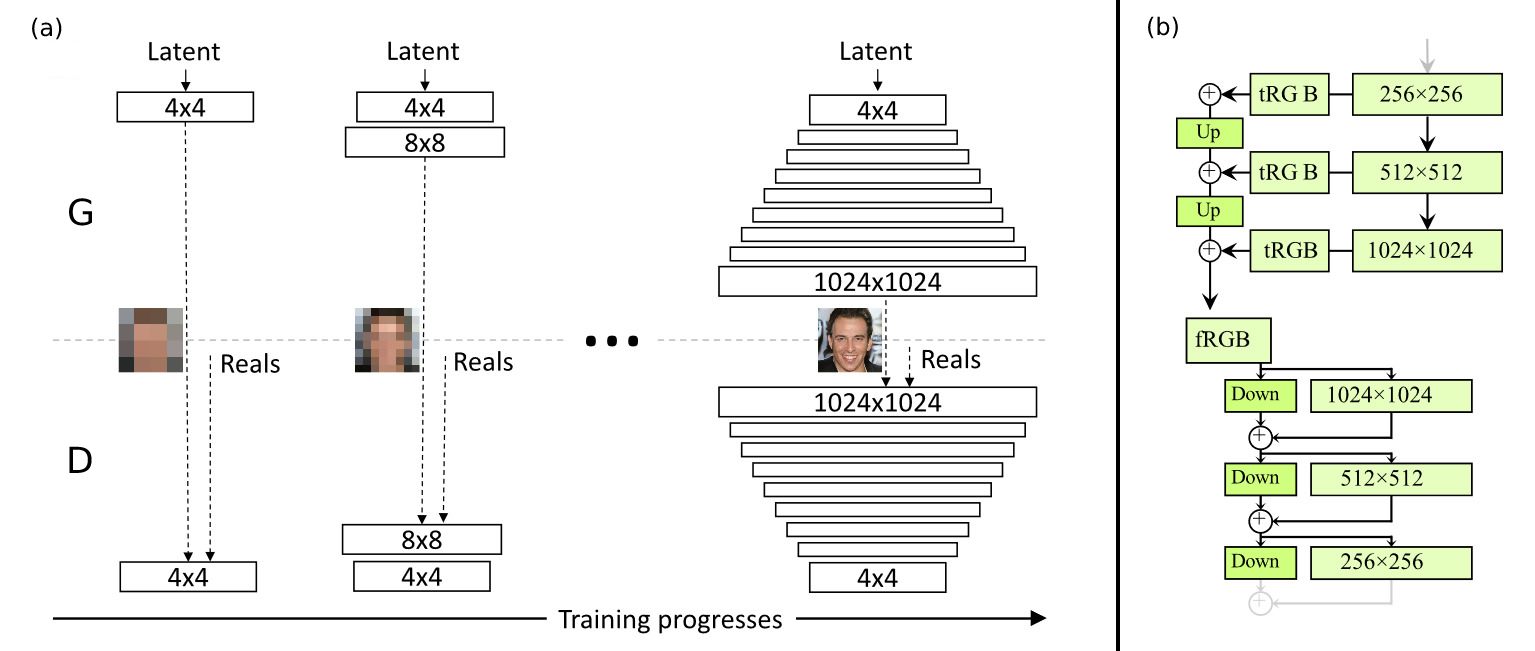}
    \vspace{-0.1in}
    \caption{(a) StyleGAN grows progressively while training, (b) StyleGAN2 does not but uses output skips and residual connections. Images from \cite{Karras2018ProgressiveGONA}, \cite{karras2020analyzing} 
    }
    \vspace{-0.1in}
    \label{fig:gan_output}
\end{figure}

\subsection{StyleGAN}

StyleGAN \cite{karras2019style} adapts the progressive training strategy from the PGGAN \cite{Karras2018ProgressiveGONA} and the generator architecture is re-designed motivated by the style transfer ideas \cite{styletransfer}. As a result, it offers a high degree of flexibility to mix image styles at different levels of its generator architecture.

StyleGAN no longer passes the random sample $z$ (often referred to as \textit{latent code} for being ``decoded'' by the generator) directly into the generator's input layer. Instead, StyleGAN starts generating images from a learned constant ($4 \times 4 \times 512$), and the latent code $z \in \mathcal{Z}$ is fed into the network along a different route (see Figure \ref{fig:gan_architectures}). First,  $z \in \mathcal{Z}$ is mapped through a deep network of fully connected layers into an intermediate latent space $w \in \mathcal{W}$. The benefit of this $\mathcal{Z}$ to $\mathcal{W}$ transformation is that the intermediate space $\mathcal{W}$ does not need to follow the Gaussian distribution of the training data (however, $\mathcal{Z}$ does). Thus, latent space $\mathcal{W}$ can be disentangled which is a desirable property because it means that features in the generated images can be controlled independently of each other, \denisnew{and this is one of the most important features of StyleGAN, as it opens up great scope for working with images in latent space}. Subsequently, for each generator layer separately, $w \in \mathcal{W}$ is converted using affine transformation (fully connected layer without activation function) into a vector of style parameters that are used to shift and scale the activity pattern in the feature maps of the respective convolutional layer. This affine transformation of feature maps is called an adaptive instance normalization (AdaIN)\cite{styletransfer}. 

This layer-wise feeding of style parameters $w \in \mathcal{W}$ allows style-mixing by feeding code parameters $w_A$ and $w_B$ of two sources A and B respectively into different layer subsets of the StyleGAN generator (see Figure \ref{fig:style_mixing}). If a code is injected into early layers it affects rough features (e.g., shape of a face) while injection into later layers correspond to finer details (e.g., skin color), \denisnew{so latent codes enable modifications at different granularities}. Finally, to provide stochastic detail that would have to be learned otherwise, pixel-wise noise is injected after each convolution. This allows the network locally stochastic placing of fine structure, such as pores, hairs, or freckles. \denisnew{All these architectural innovations allow it to outperform the previous ProgressiveGAN.}

\begin{figure}[!t]
    \centering
    \includegraphics[width=2.7in]{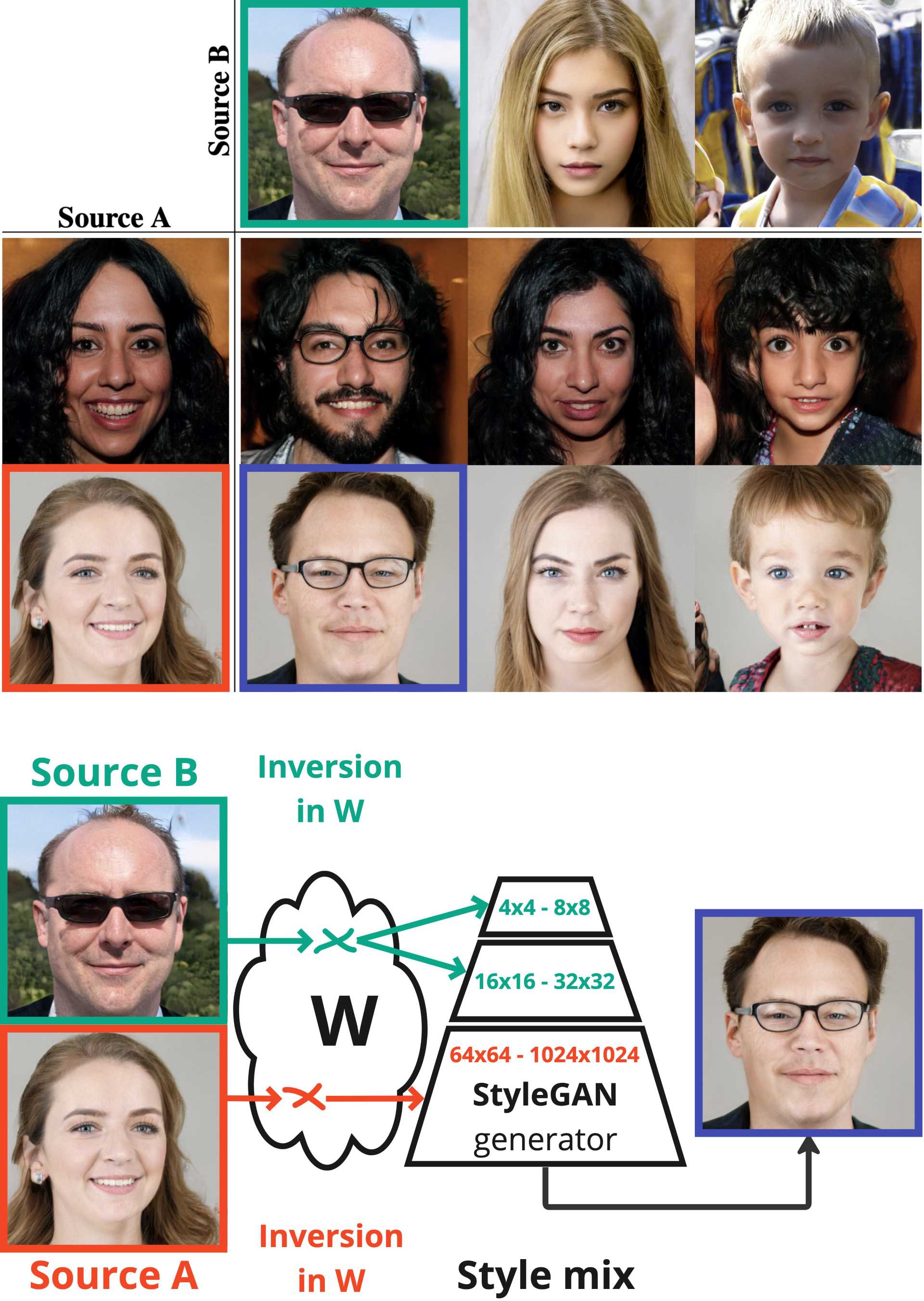}
    \vspace{-0.1in}
    \caption{Style mixing in StyleGAN\cite{karras2019style}. Top panel shows examples of style mixing, bottom panel illustrates the style mixing pipeline in StyleGAN - latent representations of two images can be used in different levels of the generator.}
    \label{fig:style_mixing}
    \vspace{-0.1in}
\end{figure}

\subsection{StyleGAN2}
StyleGAN was a major breakthrough towards the generation of high-resolution face images that looked very natural. Yet, the generated images tended to contain minor, but systematic artifacts, such as blobs or droplets. A careful analysis of this phenomenon enabled the development of an updated version, StyleGAN2 \cite{karras2020analyzing}. 

Its key change was a simplification and reorganization of the layer normalization, which in the original StyleGAN was recognized as destroying information in the relative activation strengths of the different feature maps within a layer. This was avoided by replacing the former AdaIN operations \cite{styletransfer} by a direct rescaling of the convolutional weights, again based on the the style parameter output associated with  $w \in \mathcal{W}$, followed by a normalization by the standard deviation over the scaled weights. Additionally, the noise and bias now became added outside the style block and removed from the initially learned constant input.

Another problem was that in the images created by StyleGAN some details like eye or teeth orientation seemed to be either stuck in place or jumping between positions instead of moving smoothly. This was
attributed to the generator needing to produce output images at each resolution, which forces it to generate maximal frequency details. To overcome this problem, StyleGAN2  no longer trains models using progressive growing, but sums the output from different resolutions together and utilizes skip connections (see Figure \ref{fig:gan_output}B). The size of the model itself was also increased through the number of feature maps in the layers responsible for the highest resolutions.

Furthermore, a new regularization loss was introduced to control the perceptual path length (PPL), which quantifies the smoothness of the mapping from a latent space to the output image by measuring average LPIPS  (see chapter \ref{sim:lpips}) between generated images under small perturbations in latent space.

\subsection{StyleGAN3}

While StyleGAN2 abolished the above artifacts and further increased image quality by a number of measures, a remaining problem was that, \denisnew{when making an animation footage for a specific person by manipulating its latent representation, for example rotating head or adding smile, fine details, for example hair texture, appeared to be stuck to specific image coordinates instead of properly co-moving with the object surfaces to which they were attached.}
\denisnew{This problem is demonstrated on the animation footages that can be found here \cite{StyleGAN3Page}}. It turned out that StyleGAN2 had a strong propensity to fix a feature to image coordinates whenever any information about such coordinates was available to the network.

Major sources of such information turned out to be image borders and aliasing patterns from discretization on a pixel grid. The first problem was solved by sufficient zero padding around the image in the generator \cite{xu2021positional}. The discretization problem was solved by reformulating all operations for a continuous image
, which then could be used as an equivalent representation. Correct maintenance of this equivalence required a careful enclosing of the application of the non-linearity between upsampling and downsampling operations designed to filter away any spectral out-of-band contents otherwise introduced by the non-linearity. Further changes introduced in the StyleGAN3 architecture \cite{karras2021alias} included an optimized reduction of the number of layers and simplifications, including retracting some regularizations that were introduced in StyleGAN2 but incompatible with a strict enforcement of translation invariance.
}

\maxim{Transitional and rotational equivariance is achieved by replacing the learned $4 \times 4 \times 512$ constant used in StyleGAN2 with randomly generated and information-wise equivalent Fourier features with dimensionality $36 \times 36 \times 1024$ as the first layer. To prevent leakage of absolute image coordinates into the internal representations, a  fixed-size margin around the target canvas, that is cropped after each layer, is introduced to replace the previously used padding. StyleGAN3  also uses lower cutoff frequencies for filters during up-sampling and downsampling operations to eliminate aliasing artifacts. In addition, rotational equivariance is achieved by replacing $3 \times 3$ convolutions with $1 \times 1$ convolutions, and replacing Cartesian downsampling filters with radially symmetric ones for all layers except the last two.}

\section{Measuring Similarity of Faces}
\label{ch:sim}

\ndrw{
This chapter delves into several loss functions that are applicable in evaluating the generated images and addresses the challenge of computational measures for human perception of facial images.}

\subsection{Adversarial Loss}
\label{sim:adversarial}

 \maxim{Adversarial loss is fundamental to constructing generative adversarial models \cite{goodfellow2020generative}.
The generator network $G$ learns to create samples from a given distribution, and the discriminator network $D$ learns to determine if a sample is from the real data distribution or not.
Given an image $G(z)$ generated from $z$ in a known distribution $p_Z$, and $D(x)$ being the probability of $x$ being drawn from the given data distribution $p_{data}$, the training objective of $D$ is to discriminate real images from generated ones. In other words, it needs to maximize $D(x)$ when $x$ is sampled from $p_{data}$. However, when $x=G(z)$ is produced by the generator, the generator wants $D(G(z))$ to be maximized instead.
 
 Switching to $\log(1-D(G(z))$ reverses the optimization direction for both $D$ and $G$, allowing to express the adversarial training of $D$ and $G$ as the optimization of a two-player minimax game with value function $V(G,D)$ \cite{goodfellow2020generative}:

\begin{equation}
  \begin{aligned}
    \label{eqn:AdvLoss}
    \underset{G}{\min} \underset{D}{\max} \, V(G, D) = \mathbb{E}_{x \sim p_{data}(x)} \left[\log D(x)\right] + \\ \mathbb{E}_{z \sim p_{z}(z)}\left[\log (1 - D(G(z)))\right]
\end{aligned}
\end{equation}

Although the minimax loss is commonly used to train GANs, it can lead to instability in the training process. To address this, other loss functions have been proposed, such as the Wasserstein GAN loss \cite{arjovsky2017wasserstein}, Hinge loss \cite{lim2017geometric}, and non-saturating loss \cite{goodfellow2020generative}, which have been shown to improve the stability of GANs training in various scenarios.}

\subsection{$L_2$ Loss}
\label{sim:l2}

The $L_2$ loss measures pixel-wise similarity between two images $x$ and $\hat{x}$. It is defined as $L_2(x,\hat{x}) = \sum_i{(x_i - \hat{x}_i)^2}$. While this is a fairly simple way to compute the similarity between two images, it suffers from the drawback that such pixel-wise comparison is insensitive to the local context of image points. This can make the $L_2$ distance very different from human image similarity judgements \cite{blau2018perception}. For example, for an image with a white horizontal line on a black background. If we shift the horizontal line just by one pixel down, the $L_2$ distance will be large, but the two images might be indistinguishable to a human.

\subsection{LPIPS Loss}
\label{sim:lpips}

\ndrw{It is believed that CNN features provide a more abstract representation of an image. Consequently, the distances in this representation space more accurately capture the distinctions that matter in distinguishing images from one another, while being less susceptible to low-level perturbations that are unrelated to the image's content\cite{NIPS2016_371bce7d}.}
\maxim{Thus, one approach is to transform images into a CNN feature space where point-wise distances more accurately reflect human similarity judgments, and then use the $L_2$ measure there.}
This is the underlying idea of the Learned Perceptual Image Patch Similarity (LPIPS) \cite{Zhang_2018_CVPR} measure, using CNNs trained for visual recognition tasks such as VGG\cite{simonyan2015deepVGG} or AlexNet\cite{krizhevsky2017imagenet} to transform the images. The resulting activations in CNN layers represent increasingly abstract image features, allowing the similarity measure between an image pair $x, \hat{x}$ as a weighted sum of $L_2$ distances between the corresponding feature map activations $y^l, \hat{y}^l$ with shape $(H_l, W_l, C_l)$ in the CNN layers $l$ (see eq. \ref{eqn:LPIPS}). The weighting coefficients, represented by $c_l$, are used to scale the activations channelwise and fine-tune the metric to match human similarity judgments as closely as possible, in addition to the DNN training. The LPIPS metric thus takes the form:
\begin{equation}
    \label{eqn:LPIPS}
    d(x, \hat{x}) = \sum_l \frac{1}{H_l W_l} \sum_{h, w} { \|c_l \odot (y^l_{hw} - \hat{y}_{hw}^l)\|^2_2}
\end{equation}

\subsection{Identity Preservation Loss} \label{sec:identity_preservation}

\ndrw{The losses mentioned earlier measure the overall dissimilarity between two images. However, even a minor variation in facial characteristics can lead to individuals being perceived as visually distinct (see Figure \ref{fig:ImageInversionExample} upper row vs. lower row). This, in turn, prompts the need for similarity measures designed to cater specifically to facial recognition, fine-tuned to accurately determine whether two faces correspond to the same person or not, while disregarding extraneous factors like facial position or expression.}

\maxim{To achieve this, models have been trained to enforce higher embedding similarity for intra-class face samples and larger embedding distances for cross-class samples, resulting in suitable identity preservation loss functions. The ArcFace model\cite{deng2019arcface} is a prominent example of such a model, designed specifically for face recognition tasks. The embeddings of the same face produced by this model will be close to each other, but far from the embeddings of other faces.} The loss for comparing two faces using this model is computed as $
    \mathcal{L}_{id} = 1 - \langle R(x), R(y)\rangle
$, where $R$ is the ArcFace model which produces embeddings, $x, y$ are the face images and $\langle\cdot,\cdot\rangle$ is the cosine similarity.

\subsection{Fr\'echet Inception Distance FID \label{sec:FID}}
\label{sim:metrics}

\maxim{The Fr\'echet Inception Distance (FID) \cite{heusel2017gans}  is a commonly used metric for evaluating how well the distribution of images generated by a generator matches the distribution of images used in training. FID involves embedding each image from the two distributions into a 2048-dimensional vector using InceptionV3 \cite{szegedy2016rethinking}, and then comparing the two distributions of embedding vectors using Wasserstein-2 distance.}

\maxim{The FID approach is similar to LPIPS (see Chapter \ref{sim:lpips}), because it compares higher-level features rather than RGB pixel information, both of which approximate relevant features within the human visual system. A low FID score is a good indicator that the generator is producing images that are similar to the training images.}

\maxim{FID is mostly used for measuring generator performance, not as a loss function, because it is computationally expensive. It requires a large number of images to compute covariance between all pairs of images, and passing gradients through InceptionV3 for every image. However, computing FID once to measure the perceptual quality of generated images by a trained generator is feasible and widely used.}

\section{Latent Spaces of StyleGAN}
\label{chapter:latentspace}

\ndrww{
In the original GAN architectures, the latent code was
found to be highly entangled 
and difficult to use for controlling the output image features \cite{zhu2019lia}. 
}

\begin{figure}[!t]
    \centering
    \includegraphics[width=3.5in]{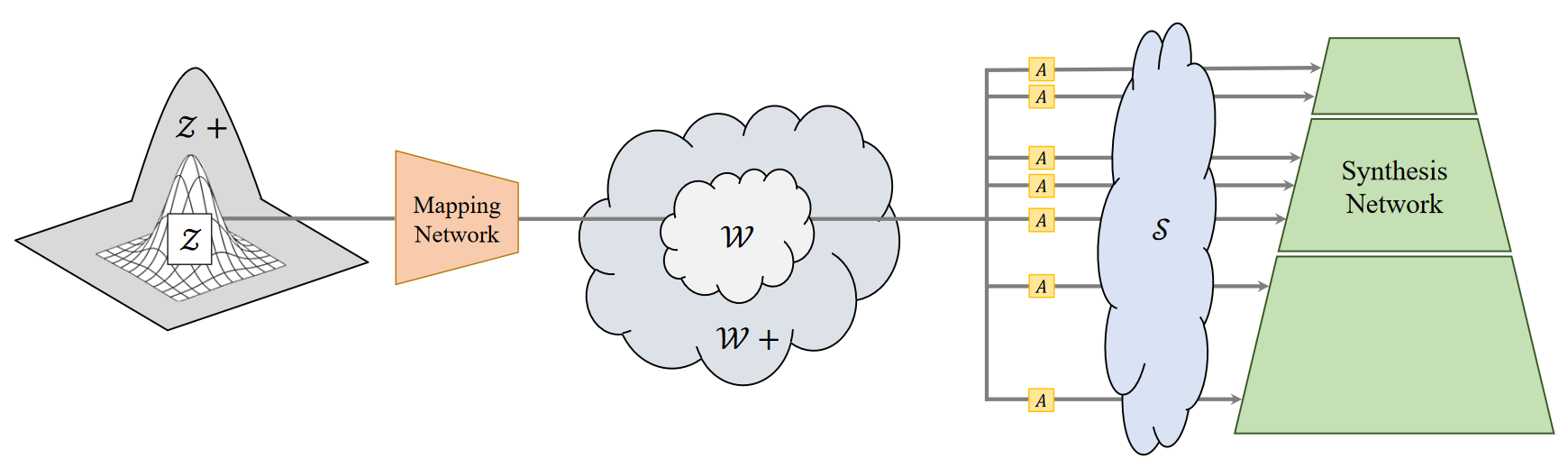}
    \vspace{-0.2in}
    \caption{Latent spaces in StyleGAN. Image from \cite{bermano2022state}.}
    \vspace{-0.2in}
    \label{fig:gan_z_w_s}
\end{figure}

\ndrw{

The first key idea of StyleGAN architectures\cite{karras2019style}\cite{karras2020analyzing}\cite{karras2021alias} is to introduces more than one innate latent space ($\mathcal{Z}, \mathcal{W}, \mathcal{S}$; see Fig. \ref{fig:gan_z_w_s}), thereby allowing to learn intermediate latent representations with properties better tailored to the semantic structure of the image space (see Fig. \ref{fig:disentanglement}). Moreover, to increase the expressive power of StyleGAN, it is common to work with extensions of these spaces ($\mathcal{Z+}, \mathcal{W+}$; see Fig. \ref{fig:gan_z_w_s}). Here, we review the commonly used spaces and describe the differences between them.

The second key idea in StyleGAN architectures\cite{karras2019style}\cite{karras2020analyzing}\cite{karras2021alias} is to inject latent codes into every layer of the generator pipeline, not just at the beginning. This allows StyleGANs to offer very flexible control modulation of the activities passing through their respective layers of the generation of images at different resolutions (see Figure \ref{fig:gan_architectures}A-C).
}

\subsection{$\mathcal{Z}$ and $\mathcal{Z+}$ spaces}
In the StyleGAN architectures, 512-dimensional samples from an isotropic normal distribution with unit variance and zero mean provide the random inputs $z \in \mathcal{Z}$ at the root of the entire generation pipeline.
$\mathcal{Z+}$ space implies sequential mapping of 18 $z \in \mathcal{Z}$ vectors into 18 corresponding $w \in \mathcal{W}$ vectors (relevant for StyleGAN and StyleGAN2 architectures with 18 layers).

\subsection{$\mathcal{W}$ space}
\label{W-space}

\begin{figure}[!t]
    \centering
    \includegraphics[width=2.8in]{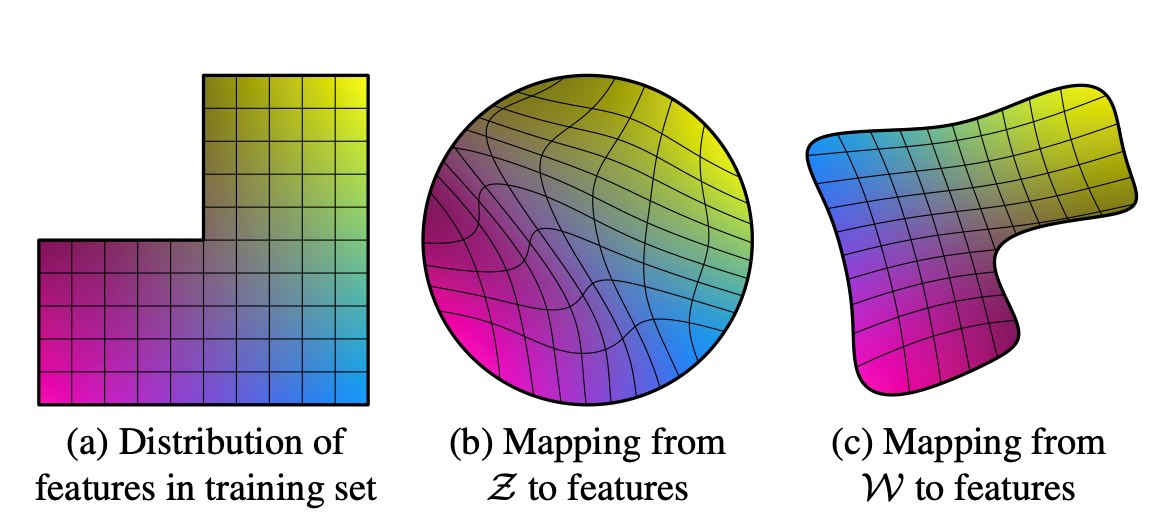}
    \vspace{-0.1in}
    \caption{The $\mathcal{W}$ space (c) demonstrates semantic axes that facilitate and generalize editing operations on different faces. To illustrate, consider the \textit{facial-beard axis} within the $\mathcal{W}$ space; when we select a point in $\mathcal{W}$ representing a face and shift it along the \textit{facial-beard axis}, the result is the appearance of facial beard on that face. In contrast, the $\mathcal{Z}$ space (b) exhibits distinct \textit{facial-beard curves} for different faces, making it less suitable for semantic editing operations across various faces. Additionally, it is important to consider that in the training dataset (a) there are examples of both male and female faces without beards, but the majority of faces with beards are male. See Chapter \ref{W-space}. Image from \cite{karras2019style}.}
    \vspace{-0.2in}
    \label{fig:disentanglement}
\end{figure}

\ndrww{
Latent codes from $\mathcal{Z}$ are transformed to latent codes in the 512-dimensional $\mathcal{W}$ space of StyleGAN through the mapping network (see Figure \ref{fig:gan_z_w_s}). 
This transformation, which must be learned during training, allows to distort the simple $\mathcal{Z}$-distribution into a distribution $\mathcal{W}$ of the same dimensionality of style parameters. In this $\mathcal{W}$ space meaningful editing operations on images can become realizable by simple axis-parallel movements of a point \cite{karras2019style}. The eight fully connected mapping layers of the StyleGAN's mapping network can provide the adaptivity to unfold the disc-shaped $\mathcal{Z}$ space into a space $\mathcal{W}$ whose shape is much closer to the required feature distribution (Figure \ref{fig:disentanglement}C, right).
Figure \ref{fig:disentanglement} illustrates a (toy) situation where the feature distribution of to-be-generated images excludes a combination of two features, leading to a distribution where one quadrant of all possible combinations is absent (Figure \ref{fig:disentanglement}A, left). To create such a distribution from the disc-shaped input distribution $\mathcal{Z}$ (Figure \ref{fig:disentanglement}B, middle) requires a very non-linear mapping. 
}

\subsection{$\mathcal{W+}$ space}
Usually, a 512-dimensional vector $w \in \mathcal{W}$ is used 18 times as the style input to 18 layers of the StyleGAN2 generator. This suggests that each of these 18 can be individually modified  for fine-tuning of a generated image. This extends latent space into 18 copies of $\mathcal{W}$ ($d = 18 \times 512$) and is denoted by $\mathcal{W+}$ (see Figure \ref{fig:gan_z_w_s}). This larger space is able to provide a different latent code for each layer of the StyleGAN generator (e.g., 18 for a StyleGAN2 generator with a 1024×1024 output resolution).

Since the StyleGAN architecture is trained using $\mathcal{W}$ space, images sampled from $\mathcal{W+}$ do not necessarily have realistic perceptual quality. 
This can allow to generate entirely novel patterns that are still face-like (e.g. "aliens"). However, it may also lead to patterns with useless structure or level of quality. As the distribution of $\mathcal{W}$ cannot be explicitly modeled, keeping the latent code in within a range that corresponds to semantically and quality-wise useful patterns is a challenging task. To learn more about trade-offs between $\mathcal{W}$ and $\mathcal{W+}$ spaces see Chapter \ref{inversion:tradeoff}.

\subsection{$\mathcal{S}$ space}
\label{ch:stylespace}

In a further step of StyleGAN processing, the latent code $w \in \mathcal{W}$ of an image is further transformed to $s \in \mathcal{S}$ vectors for each layer of the StyleGAN generators \cite{Wu2021StyleSpaceAD}. The details of these transformations differ slightly between the versions, but a shared commonality is a mapping to a parameter vector of style parameters
$\mathcal{S}$ that parametrizes a set of affine transformations (one for each layer) that either normalize the activity pattern in a layer (in the case of StyleGAN), or that directly define a two-step scaling of the weights of a layer (mod/demod operations of StyleGAN2 and StyleGAN3). While the activity normalization in StyleGAN requires a specification of two parameters (bias and scaling) for each feature map, StyleGAN2/3 get by with a mere scaling (single parameter) for the feature map scalings in the mapping layers. For the sake of brevity, we focus on the case of StyleGAN2 \cite{karras2020analyzing} in the following discussion, which is very representative of the major ideas behind the style mapping. 

At the style input of every layer of the StyleGAN2 generator is an independent single-layer perceptron 
denoted by A (affine transformation). This network maps $w \in \mathcal{W}$ vector into a new vector $s \in \mathcal{S}$ vector that provides for each of the layer's weight kernel a separate scalar scaling parameter. The size of the vector $s \in \mathcal{S}$ equals  the number of channels in all layers of the StyleGAN2 generator. For example, in StyleGAN2 $\dim(\mathcal{W}) = 512$ and $\dim(\mathcal{S}) = 9088$ \cite{Wu2021StyleSpaceAD}.
This modulated kernel is then applied to the layer $L-1$ of the StyleGAN2 generator to produce the activation of the channel in the layer $L$  of StyleGAN2. In \cite{Wu2021StyleSpaceAD} Wu et al.  proposed to name this latent space of coefficients $s$ StyleSpace $\mathcal{S}$. Analysis from Wu et al. indicates, that the $\mathcal{S}$ space is more semantically disentangled than previous latent spaces of StyleGAN2. The usecases  of $S$ space for face editing is described in Chapters \ref{ch:StyleSpace} and \ref{ch:StyleCLIPglobal}.

\section{Inversion to Latent Spaces of StyleGAN} \label{ch:Inversion}

This chapter will 
review recently developed solutions 
that map an image to a suitable StyleGAN latent code.
It is important to know the latent representation of an image of a face in StyleGAN space in order to manipulate that image or combine it with the style of some other image of a face. 
Since a StyleGAN model has several latent spaces (see Figure \ref{fig:gan_z_w_s}), this task can come in different variants, depending on the choice of the latent space for which a representation is sought. Furthermore, we shall see that these different choices may entail different properties, e.g. how the image will change under manipulations of its latent code, but also how well the inversion can be steered to faithfully capture the important visual features of the given real image. 

After a brief overview over the major groups of inversion methods in general in Chapter \ref{inversion:overview} we turn in Chapter \ref{inversion:evaluating} to the important issue of \textit{evaluating} the inversion for distortion, perceptual quality, and editability, and how to control a suitable trade-off between these properties. Achieving an approximate resemblance is relatively easy (see Figure \ref{fig:ImageInversionExample}), but resemblance in fine detail is very important for the perception of faces.
Then, in Chapter \ref{inversion:encoders}, we delve into specific methods for obtaining inversion encoders that can solve the inversion task: \textit{pixel2style2pixel} (pSp) \cite{richardson2021encoding}, \textit{encoder4editing} (e4e) \cite{Tov2021DesigningAE} and \textit{ReStyle} \cite{alaluf2021restyle}. Finally, Chapter \ref{inversion:tuning} describes techniques for improving inversion quality
by selective tuning of StyleGAN generator weights: \textit{Pivotal Tuning} \cite{roich2022pivotal} introduces optimization-based fine-tuning of StyleGAN; \textit{MyStyle} \cite{nitzan2022mystyle} extends fine-tuning to hundreds of portrait images of a given person; \textit{HyperStyle} \cite{alaluf2022hyperstyleNA} introduces encoder based prediction of fine-tuning weights of the StyleGAN generator.

\subsection{Major groups of inversion methods}\label{inversion:overview}

\begin{figure}[!t]
    \centering
    \includegraphics[width=3.5in]{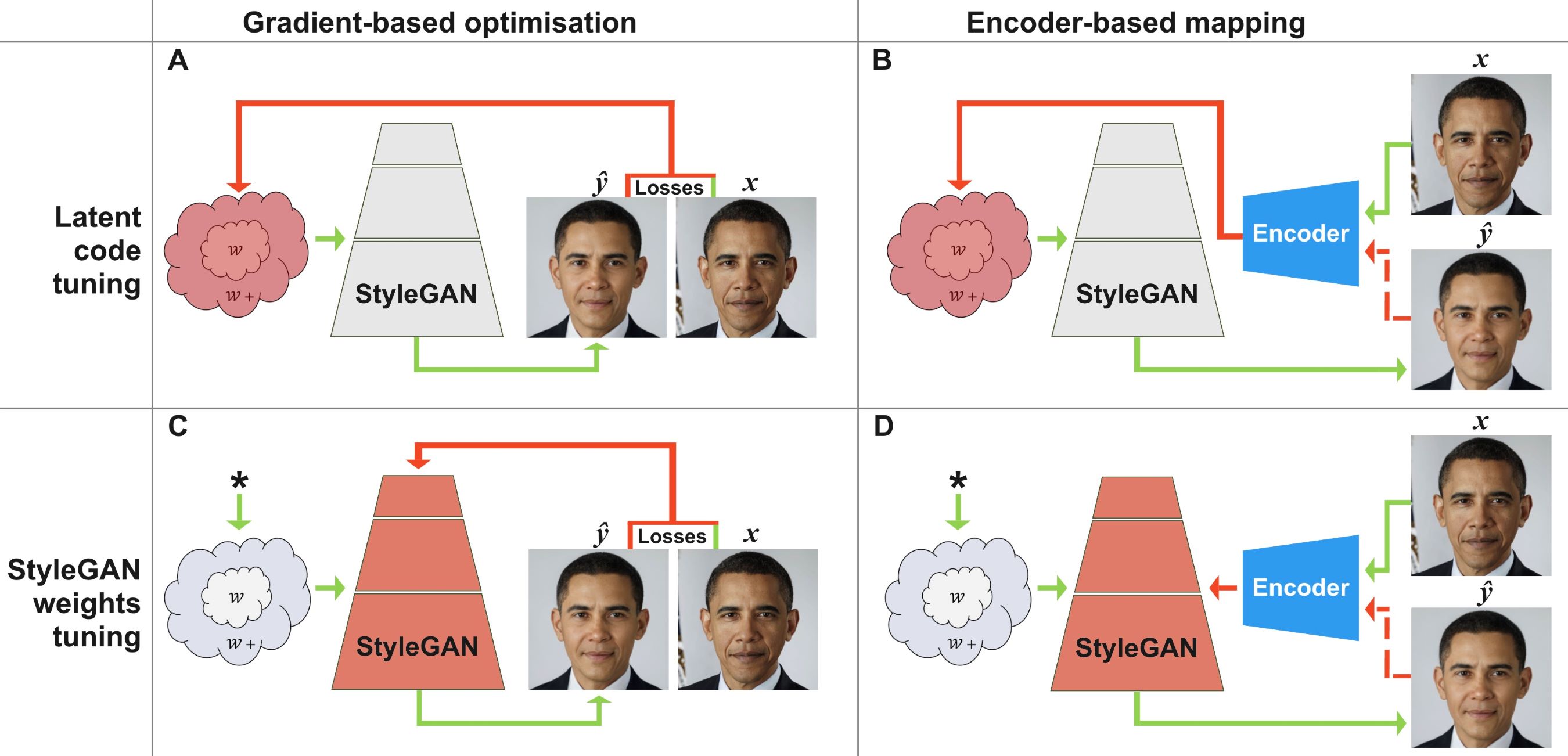}
    \caption{Inversion methods: target image $x$,  reconstruction image $\hat{y}$ , initial inversion $*$. Typical losses are $L_2$, LPIPS \cite{Zhang_2018_CVPR}, ArcFace loss \cite{deng2019arcface}.}
    \vspace{-0.2in}
    \label{fig:InversionMethods}
\end{figure}

Inversion methods can typically be divided into three major groups of methods: \textit{gradient-based optimization} of the latent code (Chapter \ref{inversion:optim-based} and Figure \ref{fig:InversionMethods}A), direct \textit{encoder-based mapping} onto the latent code (Chapter \ref{inversion:encoder-based} and Figure \ref{fig:InversionMethods}B), and \textit{fine-tuning} of weights of the StyleGAN generator (Chapter \ref{inversion:generator-tuning} and Figure \ref{fig:InversionMethods}C \& D).

\subsubsection{Gradient-Based Optimization of the Latent Code}
\label{inversion:optim-based}

Gradient-based optimization methods (see Figure \ref{fig:InversionMethods}A) \cite{abdal2020image2stylegan++}\cite{karras2020analyzing}\cite{Lipton2017PreciseRO} directly optimize the latent vector using gradients from the loss between the real image and the generated one. Such methods can find a latent representation of the original image with a reasonable similarity (see Figure \ref{fig:ImageInversionExample}). However, there still exist three main drawbacks \cite{abdal2020image2stylegan++}:
\begin{itemize}
    \maxim{
    \item Optimizing the procedure is a time-consuming task that usually takes several minutes on a modern GPU.
    \item The random initialization choice can significantly impact the final reconstruction image.
    }
    \item The found latent point in $\mathcal{W}$ or $\mathcal{W+}$ spaces through inversion optimization steps is less stable while editing of the generated image than latent points obtained by sampling from $\mathcal{Z}$ space and generating latent points in $\mathcal{W}$ or $\mathcal{W+}$ spaces through the mapping network (from $\mathcal{Z}$ to $\mathcal{W}$ space). The mapping network generates points in the distribution of the training dataset, while inversion optimization steps can move the latent point away from the distribution of the training dataset (see Chapter \ref{inversion:metrics}).
\end{itemize}

\subsubsection{Encoder-Based Mapping to the latent code}
\label{inversion:encoder-based}

\begin{figure}[!t]
    \centering
    \includegraphics[width=2.5in]{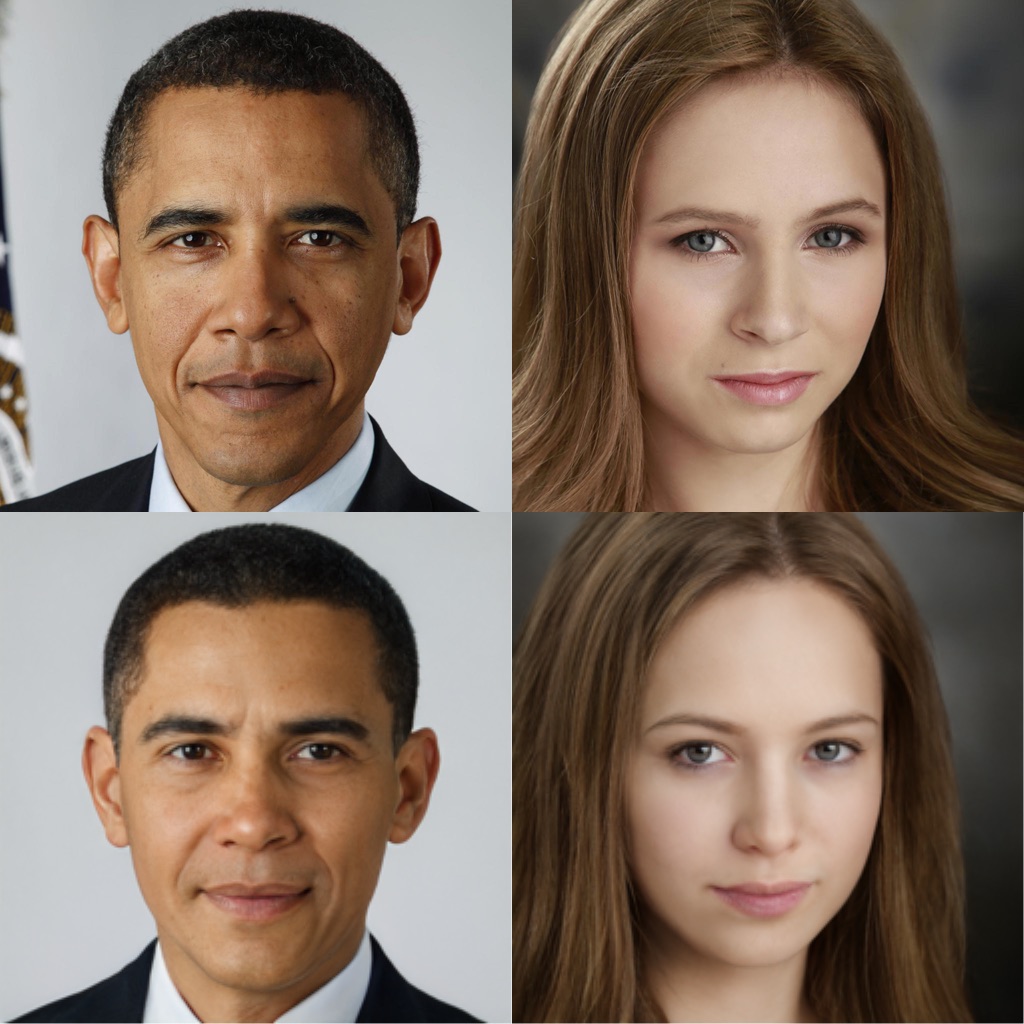}
    \caption{Obama \cite{Obama} and Actress \cite{WhiteFemale} before (upper row) and after Inversion (lower row). This inversion was performed using StyleGAN2-ada \cite{karras2020training}.}
    \vspace{-0.2in}
    \label{fig:ImageInversionExample}
\end{figure}

Alternatively, encoder-based methods for finding the latent code (see Figure \ref{fig:InversionMethods}B) \cite{richardson2021encoding}\cite{Tov2021DesigningAE} train an encoder network over a large number of samples to directly map from the RGB image space into a latent space of StyleGAN. Once trained, the encoding can be done in the fraction of a second needed to process through the CNN encoder. Latent points obtained from the encoder network are more suitable for editing \cite{Tov2021DesigningAE} by moving the point in the latent space, as the encoder network is trained to generate points inside the distribution of the training dataset of StyleGAN. However, training such an encoder is not trivial 
and
the image generated from the obtained latent code
may lose the identity of the original face (see Figure \ref{fig:ImageInversionExample}).

Conventional image- and feature-level losses (i.e. MSE and Perceptual losses \cite{Zhang_2018_CVPR}\cite{johnson2016perceptual}) between the input image and the reconstructed image, may not be enough to guide the training of the encoder network. Wei et al. \cite{Guan2020CollaborativeLF} proposed the method of training the encoder in cooperation with  an optimization-based iterator. One more possibility for getting a better inversion quality is to compute the loss based on SSIM \cite{1284395SSIM}, Identity loss \cite{deng2019arcface} and LPIPS \cite{Zhang_2018_CVPR}) to train an encoder that maps RGB into $\mathcal{W}$ space. An additional option is encoding into the $\mathcal{W+}$ latent space that provides finer control over the generator, utilizing regularization methods while training of such RGB to $\mathcal{W+}$ encoder \cite{Tov2021DesigningAE}.

\subsubsection{Fine-tuning of the StyleGAN generator}
\label{inversion:generator-tuning}

When an image of a face includes something outside the training distribution e.g., a tattoo
(see Fig. \ref{fig:PivotalTuningExamples}), it is difficult to find a good inversion, as there is no such a point in the latent space of
StyleGAN 
that allows for reconstruction of such 
details. In this case, a possible solution is to fine-tune the StyleGAN generator weights themselves, using the target image or a set of target images \cite{roich2022pivotal}. This motivates a set of methods that operate directly on the generator weights to improve the inversion quality of a given image.

Before starting such fine-tuning of the StyleGAN generator, the target image must be first inverted into StyleGAN's latent space (see Fig. \ref{fig:InversionMethods}C\&D) to the best possible reconstruction match (see Fig. \ref{fig:InversionMethods}A\&B) using the previously discussed approaches. Then the StyleGAN generator model can be fine-tuned using loss-functions applied to the reconstructed and target images (see Fig. \ref{fig:InversionMethods}C and Chapter $\ref{inversion:pti}$).

Fine-tuning the StyleGAN generator by gradient-based optimization (Fig. \ref{fig:InversionMethods}C) for each new image requires a couple of minutes of computation. This makes such methods difficult to apply in practice. By analogy with encoder-based methods for predicting the inversion latent vector (see Fig. \ref{fig:InversionMethods}B), we can, here too, seek an encoder whose output is used to modulate the weights of the StyleGAN generator (Fig. \ref{fig:InversionMethods}D). An example of such an approach is \textit{Hypernetworks} \cite{alaluf2022hyperstyleNA}. The created \textit{Hypernetwork} is trained to scale channel-weights of kernels of selected layers of the StyleGAN generator to match the target image and image generated by StyleGAN from the latent code (see Fig. \ref{fig:InversionMethods}D and Chapter \ref{inversion:hyperstyle}).

\subsection{Evaluating GAN Inversions}\label{inversion:evaluating}
\subsubsection{Reconstruction quality vs. editability in StyleGAN}
\label{inversion:metrics}

In addition to preserving similarity to the original image, the central motivation of the inversion step  is to facilitate further latent editing operations. There exist a variety of points in the latent space that result in similar images to the original one, some of these points are more suitable for latent editing than others \cite{Tov2021DesigningAE}\cite{Wei2022E2StyleIT}\cite{zhu2020domain}. A successful encoding of a real image into a latent space should enable decent editability via the latent code.

\subsubsection{Distortion, perceptual quality, editability}  %

Following the above observations, inversion methods and reconstruction quality should be evaluated based on several components: \textbf{distortion}, \textbf{perceptual quality} and \textbf{editability}. Simply put, \textbf{distortion} is a dissimilarity (e.g. MSE, SSIM, LPIPS) 
between the original and reconstructed images \cite{blau2018perception}. Distortion alone, however, does not capture the quality of the reconstruction. \textbf{Perceptual quality} measures how realistic the reconstructed images are (e.g. adversarial discriminator), with no relation to any reference image \cite{blau2018perception}. \textbf{Editability} stands for maintaining high perceptual quality of the image generated from the \textit{edited} latent code \cite{Tov2021DesigningAE}. The work \cite{blau2018perception} proved that there exists an explicit trade-off between distortion and perceptual quality. Therefore, the distortion, perceptual quality and edibility of the reconstructed images must be evaluated to provide a complete evaluation of the inversion method.

\subsubsection{Distortion vs. perceptual quality \& editability trade-off}  %
\label{inversion:tradeoff}

\begin{figure}[!t]
    \centering
    \includegraphics[width=3.0in]{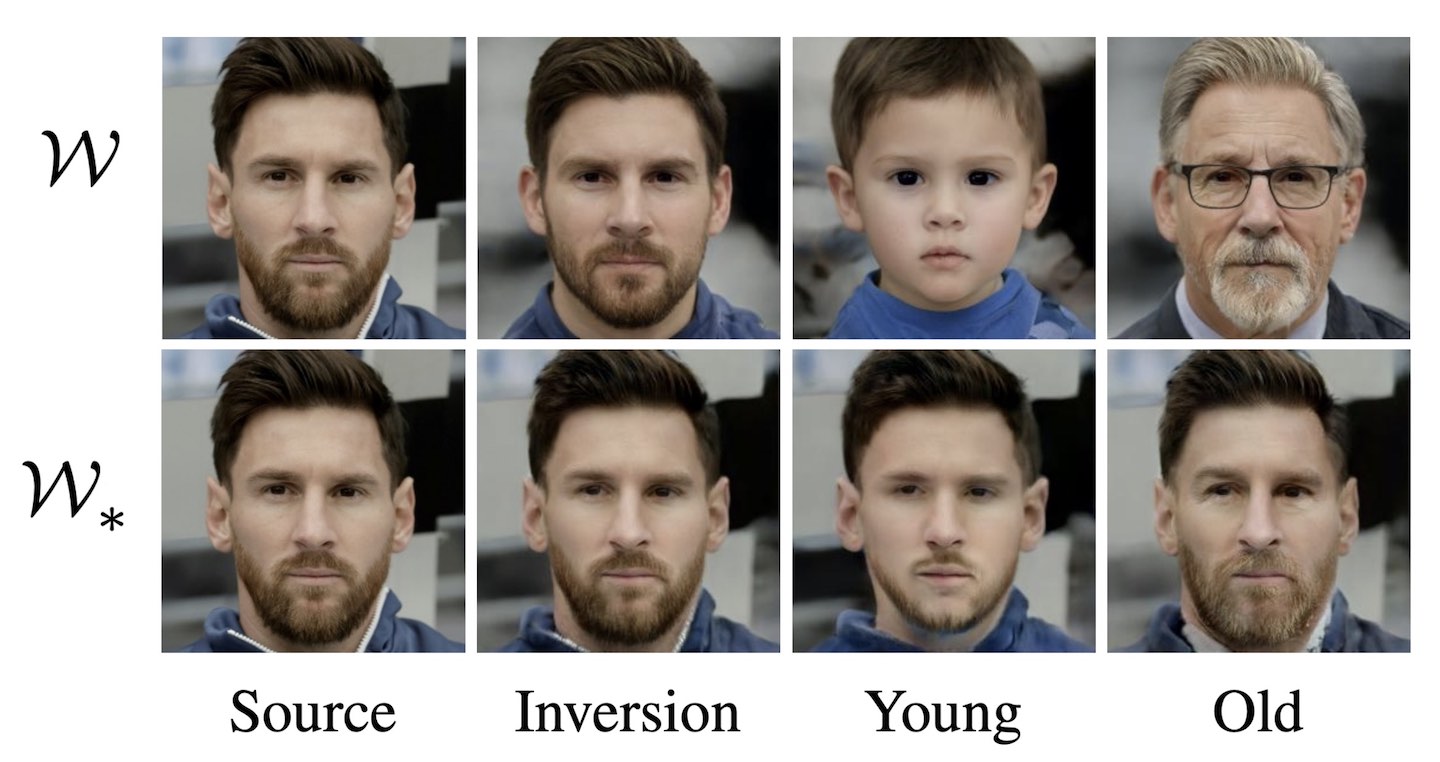}
    \caption{The editability gap in $\mathcal{W}_*$ space. In paper \cite{Tov2021DesigningAE} the space of vectors in $\mathcal{W}$ that are out of the mapping network manifold was denoted as $\mathcal{W}_*$.}
    \vspace{-0.2in}
    \label{fig:Inversion}
\end{figure}

All possible output vectors of the StyleGAN mapping network constitute its latent space $\mathcal{W}$, where the input vectors of the mapping network are normally distributed vectors in $\mathcal{Z}$ space. Latent vectors generated as the result of an inversion method may, however, not necessarily lie within this manifold of the StyleGAN mapping network $\mathcal{W}$. In
\cite{Tov2021DesigningAE} and Figure \ref{fig:Inversion} the set of such vectors is denoted as $\mathcal{W}_*$.
Moreover, manipulation via the inverted image code can be performed not only in $\mathcal{W}$ or $\mathcal{W}_*$ latent spaces of StyleGAN, but also in $\mathcal{W}+$ space, where vectors for 18 layers of StyleGAN2 are independent but each 512-dim style vector belongs to $\mathcal{W}$ space.
In $\mathcal{W}$ space, reconstructions have the highest distortion, but good editability and perceptual quality.
In $\mathcal{W}_*$ space (see Figure \ref{fig:Inversion}), reconstructions have the lowest distortion, but the worst editability and perceptual quality after editing.
In $\mathcal{W}+$ space, reconstructions have low distortion, good editability and perceptual quality.

\subsection{Inversion Encoders}
\label{inversion:encoders}
Inversion encoders solve the GAN inversion task in the form of a direct mapping  from an image onto a latent code.
Ideally, the concatenation of the GAN and the Inversion Encoder should produce the identity mapping. An early version of this idea came up in BiGAN \cite{donahue2016adversarial}, where the forward and backward mappings are developed simultaneously. However, with regard to StyleGAN inversion, the inverted part of the GAN is not the entire path from the image to its $z$-input, but to one (or several) of its intermediate latent spaces (see Chapter $\ref{chapter:latentspace}$).

\subsubsection{pSp Encoder: Image-to-StyleGAN} %
\label{inversion:psp}

\begin{figure}[!t]
    \centering
    \includegraphics[width=3.5in]{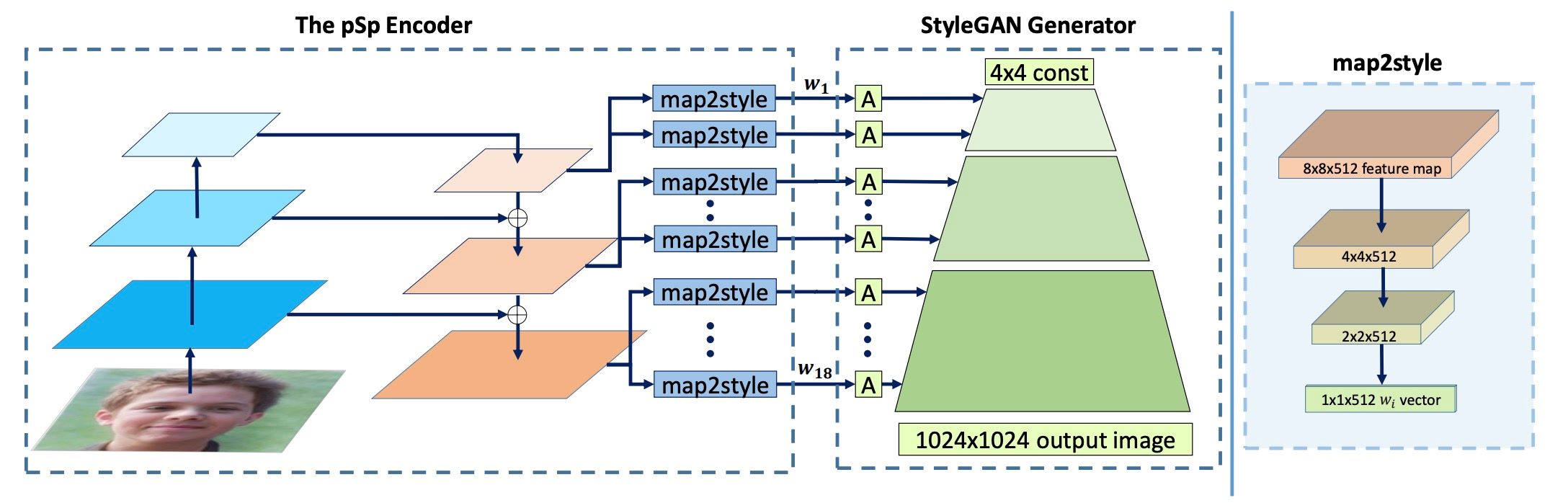}
    \vspace{-0.2in}
    \caption{pSp encoder architecture \cite{richardson2021encoding}.}
    \vspace{-0.2in}
    \label{fig:pSp}
\end{figure}

The idea behind the \textit{pixel2style2pixel} (pSp) inversion encoder into $\mathcal{W}+$ space \cite{richardson2021encoding} is based on the fact that different layers of the StyleGAN architecture correspond to different levels of generated detail (coarse, medium, and fine). Similarly, different layers of CNN-encoder feature maps also correspond to different levels of detail (coarse, medium, and fine). In the pSp encoder (see Fig. \ref{fig:pSp}) feature maps are first extracted using a standard feature pyramid over a ResNet backbone. For each of the 18 target styles, a small mapping network is trained to extract the learned styles from the corresponding feature map, where styles (0-2) are generated from the small feature map, (3-6) from the medium feature map, and (7-18) from the largest feature map. The mapping networks, \textit{map2style}, are small fully convolutional networks,
each of which generates a 512 $\Delta_i$ vector that are added to $\overline{w}$, where $\overline{w}$ is the mean of the distribution of the \textit{mapping network} output vectors when sampling in $Z$ space \cite{richardson2021encoding}. The resulting $18 \times 512$ vectors $\overline{w} + \Delta_i$ in the $W+$ space are fed into the StyleGAN, starting from its matching affine transformation, $A$ (see Fig. \ref{fig:pSp}) \cite{richardson2021encoding}. The pSp encoder is trained using the $L_2$, LPIPS \cite{Zhang_2018_CVPR}, Identity \cite{deng2019arcface} and additional regularization loss functions. For the training of the encoders the weights of the StyleGAN generator are frozen, leaving as the adapted components only the ResNet backbone, the upsampling layers, and the \textit{map2style} mapping networks.

\subsubsection{e4e Encoder for StyleGAN Image Manipulation}
\label{inversion:e4e}

The authors of \textit{encoder4editing} or simply \textit{e4e} (Fig. \ref{fig:e4e_and_ReStyle}A) \cite{Tov2021DesigningAE} approach took the above described pSp architecture \cite{richardson2021encoding} (Fig. \ref{fig:pSp}) as their starting point and proposed a novel encoding scheme. The encoder's input is an image and the encoder output is a tensor of shape $18 \times 512$ in the $W+$ space, whose first $1 \times 512$ values specifies the StyleGAN2's $w$ vector, and its remaining $17 \times 512$ values represent the $\Delta_i$ vectors for the last 17 layers of the StyleGAN2 generator. Now, instead of training all components of this tensor simultaneously, the authors propose a \textit{progressive} learning scheme that they structure in the following way: first, learning starts with setting all $17 \times 512$ $\Delta_i=0$, thus training only the first $1 \times 512$ values for matching StyleGAN2 $w$ vector. Subsequently, the mapping networks sequentially unfreeze learning of their $\Delta{i}$ outputs for the higher layers $i=2\ldots 18$ of the StyleGAN2, in ascending order. This scheme allows the encoder to first learn a coarse reconstruction, to which it then learns to add increasingly finer details.

Additionally, they propose a refined regularization scheme for the $17 \times 512$ $\Delta_i$ vectors to keep $w + \Delta_i$ in the space $\mathcal{W}$ to achieve a better editability, due to the individual adjustment of the style vector $w$ for each layer of generator, along with improved perceptual quality. This regularization scheme employs three parts: a $L_2$ regularization loss is used to minimize the joint variance of all $\Delta_i$ ($\mathcal{L}_{d-reg}$ in Figure \ref{fig:e4e_and_ReStyle}A). A second part is a loss derived from a latent discriminator ($\mathcal{L}_{adv}$ in Figure \ref{fig:e4e_and_ReStyle}A) which is trained in an adversarial manner to discriminate between latent codes from the ``true'' $\mathcal{W}$ (obtained by feeding samples from $\mathcal{Z}$ to the StyleGAN encoder), and the encoder’s learned latent codes. Finally, to make the $\Delta_i$ also to contribute to distortion reduction, additional distortion losses ($L_2$, LPIPS \cite{Zhang_2018_CVPR}, and Identity preservation ArcFace \cite{deng2019arcface}) are added.

\subsubsection{ReStyle: Iterative Inversion Refinement}
\label{inversion:restyle}

\begin{figure}[!t]
    \centering
    \includegraphics[width=3.5in]{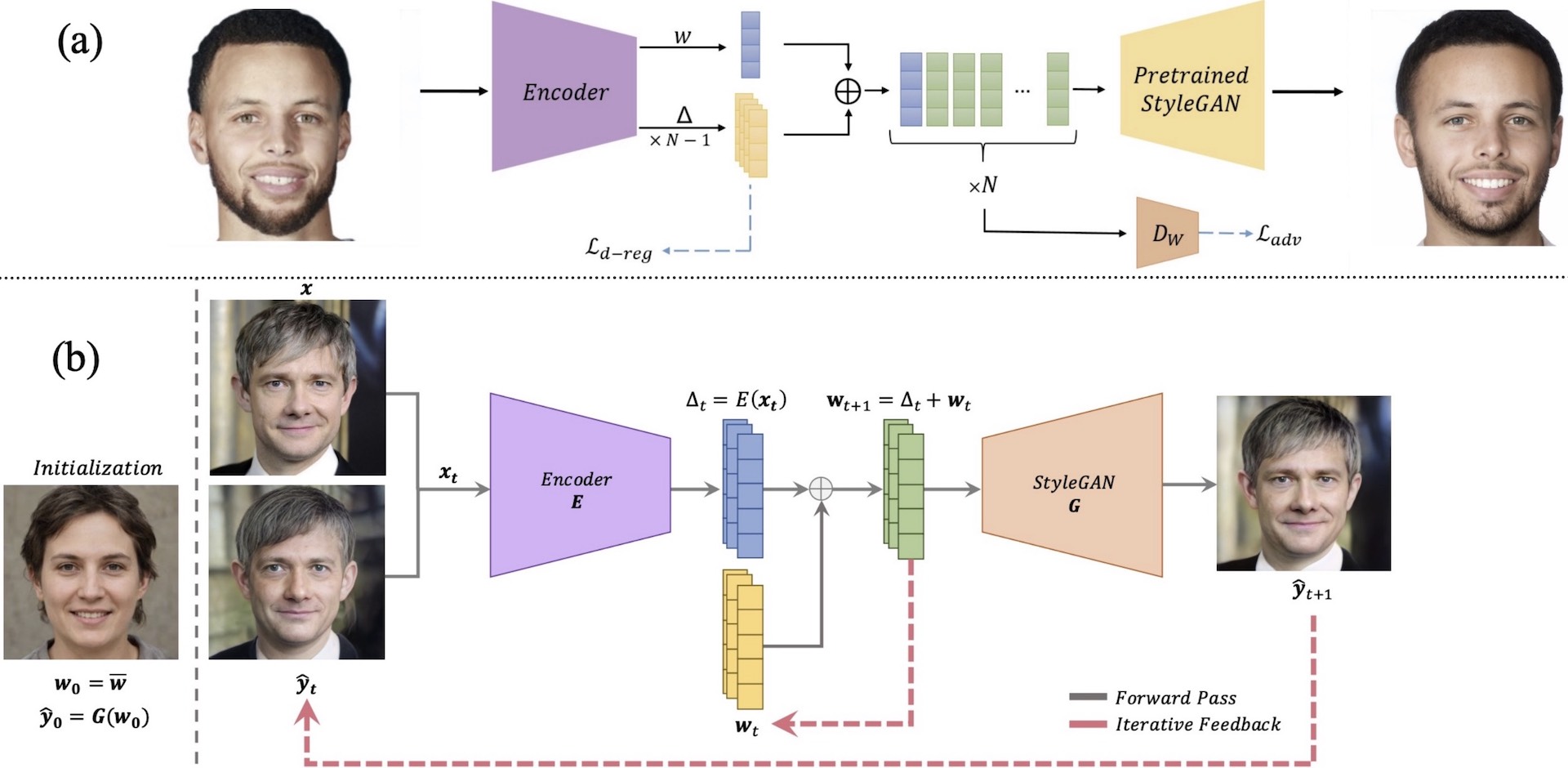}
    \caption{(a) e4e encoder architecture with \textit{progressive} training of encoder sequence by combining regularization, adversarial and distortion losses (not depicted) to improved control of editability-distortion trade-off \cite{Tov2021DesigningAE}. (b) \textit{ReStyle} iterative inversion scheme \cite{alaluf2021restyle}.}
    \vspace{-0.2in}
    \label{fig:e4e_and_ReStyle}
\end{figure}

\maxim{The \textit{ReStyle} \cite{alaluf2021restyle} method differs from typical encoder-based inversion methods, which infer the inverted latent code of the input using a single forward pass, by adding an iterative inversion mechanism with an additional feedback input. (see Figure \ref{fig:e4e_and_ReStyle}B). 
The encoder is fed with the output of the previous iteration along with the original input image, using several forward passes. This approach enables the encoder to focus on the relevant regions while leveraging the knowledge acquired in earlier iterations.}

\maxim{This method begins with an initial reconstruction $\hat{x_0}=G(w_0)$ of a source image $x$, generated from an initial representation $w_0$. To predict a sequence $w_t, t=1..N$ of image style codes, this method performs $N > 1$ steps. The final inversion $w=w_N$ and its corresponding reconstruction $\hat{x} = G(w)$ are the final results. At each step $t$, the encoder $E$ receives an input $x_t=(x,\hat{x}_t)$ consisting of the original image $x$, paired with its reconstruction $\hat{x_t}$ from the most recent time step, to compute a refined new residual code $\Delta_t=E(x_t)$, which is added to the inversion code of the source image as
$w_{t + 1} = \Delta_t + w_t$. This new latent $w_{t+1}$ is passed through the StyleGAN generator again to update image reconstruction $\hat{x}_{t+1} = G(w_{t+1})$, which is used in the next iteration. 
}

The \textit{ReStyle} approach demonstrates better $L_2$, LPIPS \cite{Zhang_2018_CVPR}, and Identity \cite{deng2019arcface}) metrics than encoder-based approaches \cite{Tov2021DesigningAE}\cite{richardson2021encoding}
delivering high-quality reconstructions while maintaining fast inference times.

\subsection{Fine-tuning of the StyleGAN generator}

\label{inversion:tuning}

\begin{figure}[!t]
    \centering
    \includegraphics[width=3.5in]{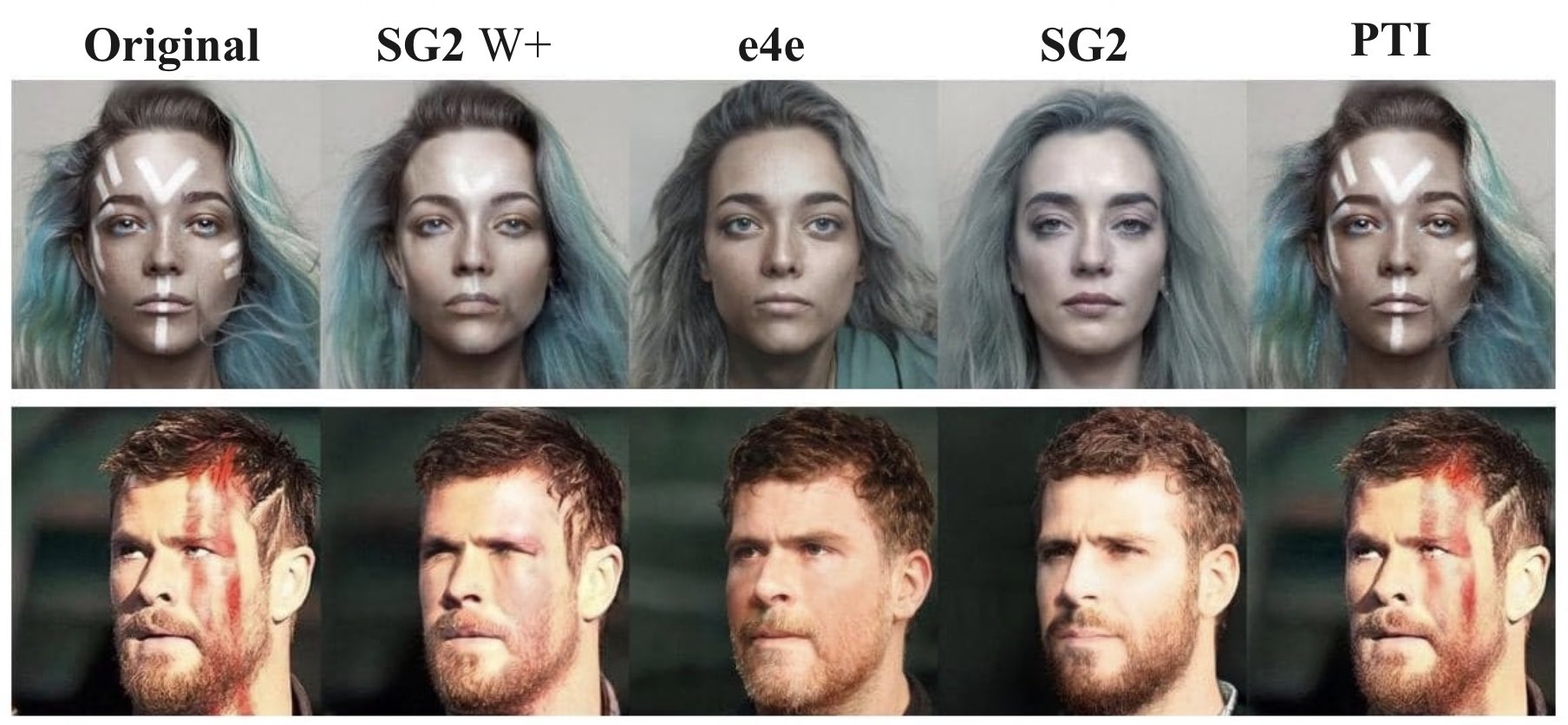}
    \caption{Comparison of \textit{Pivotal Tuning} \cite{roich2022pivotal} and other inversion methods (see \cite{karras2020analyzing} \cite{abdal2020image2stylegan++} \cite{roich2022pivotal} and Chapter \ref{inversion:e4e}))}
    \vspace{-0.2in}
    \label{fig:PivotalTuningExamples}
\end{figure}

Images of real faces often contain various unique details such as tattoos, scars, fashion elements or light. It is challenging to apply identity-preserved editing to such out-of-domain face images even with the previous methods. Inversion of such face images may lead to poor results far away from the generator's domain, because the nearest image in this domain may not have all these details. As a result they will be lost after the editing process (see Fig. \ref{fig:PivotalTuningExamples}).

\subsubsection{Latent-based Editing of Real Images}
\label{inversion:pti}

To better cope with the inversion problem of unique face details (e.g., tattoos) the \textit{Pivotal Tuning} \cite{roich2022pivotal} method adds a tuning step for the StyleGAN generator. The method inverts the source image $x$ to style code $w_p \in \mathcal{W}$ in the native latent space of StyleGAN, which will be called as the \textit{pivot} code. Let $x^p = G(w_p; \theta)$ be the image generated from the \textit{pivot} code $w_p$ when using the generator with weights $\theta$. Then, the tuning step consists of adapting the weights between the \textit{pivot} code and the output to bring $x^p$ closer to $x$ using LPIPS and $L_2$ losses. The mapping network layers between $z$ and $w$ remain frozen. However, it is important to constrain the changes of the tuning such that they only affect the reconstruction mapping within a neighborhood of the \textit{pivot} code $w_p$. To achieve this, they introduce a suitably designed regularization term, implemented in the form of an iterative process. At each iteration, a $z$ value is sampled from the standard normal distribution and the StyleGAN mapping network produces corresponding style code $w_z$. The difference $w_z-w_p$ and $w_p$ gives a direction away from the \textit{pivot} point whose length is scaled to the value of a coefficient $\alpha$. The resulting end point $w_r$ of this scaled vector, attached to the \textit{pivot} point $w_p$ (see equation \ref{eqn:PTIRegularization}) is a code in a certain neighborhood of the \textit{pivot} $w_p$:
\begin{equation}
    \label{eqn:PTIRegularization}
    w_r = w_p + \alpha \frac{w_z - w_p}{\|w_z - w_p\|_2}.
\end{equation}

It contributes a regularization step for the fine-tuned StyleGAN generator by adapting the weights $\theta^*$ of the latter towards minimizing the distance between the image pair $(x_r,x_r^*)$ obtained from $w_r$ with original generator ($x_r = G(w_r, \theta)$) and the tuned generator ($x_r^* = G(w_r, \theta^*)$), using LPIPS and $L_2$ loss functions. This neighborhood-restricted tuning process ends up altering appearance features that represent mostly face identity, without interfering with editing capabilities.

\subsubsection{MyStyle: A Personalized Generative Prior}
\label{inversion:myStyle}

\begin{figure}[!t]
    \centering
    \includegraphics[width=2.5in]{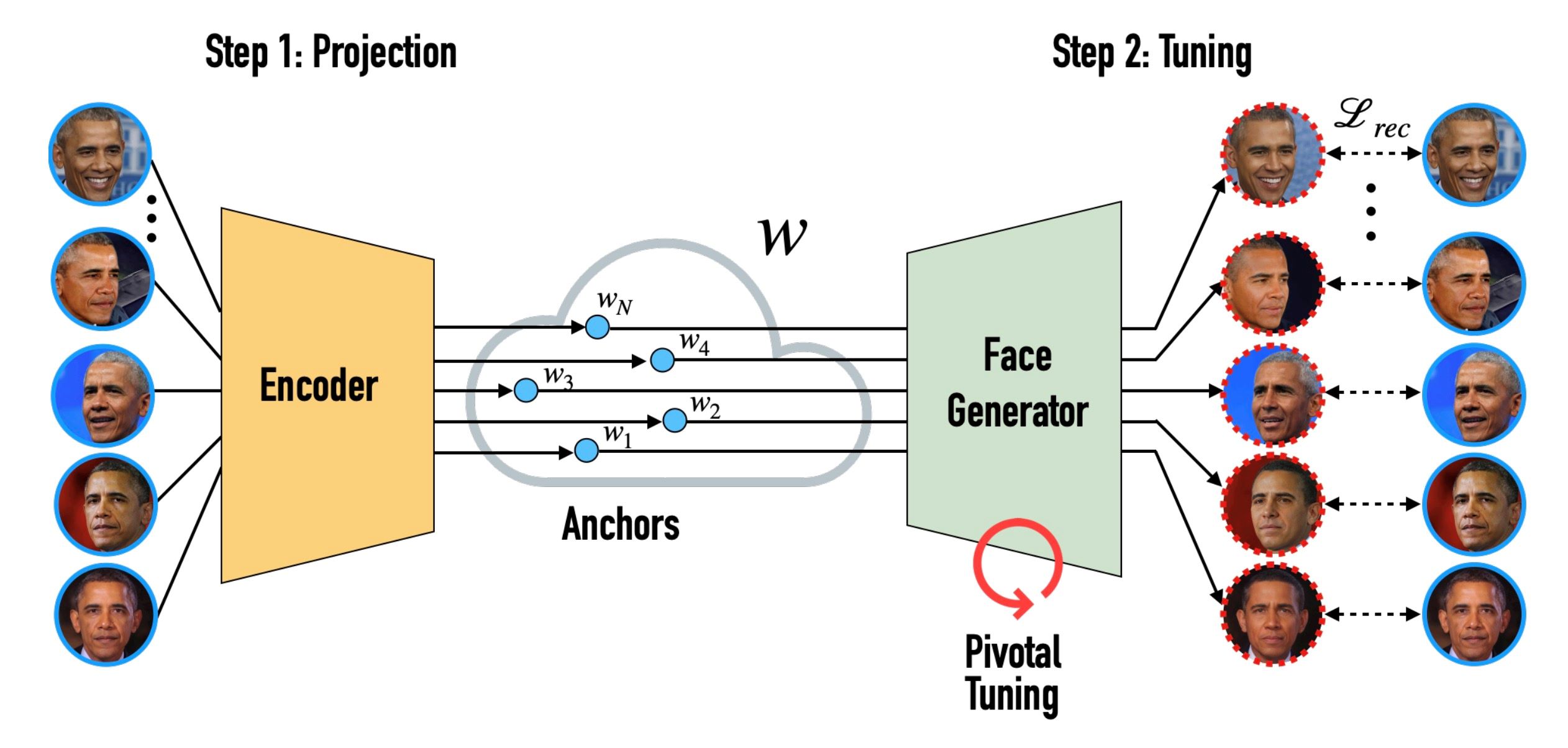}
    \caption{\textit{MyStyle} fine-tuning scheme of the StyleGAN generator \cite{nitzan2022mystyle} using \textit{pivotal tuning} around a large number of \textit{anchors} given by a collection of views of the face of a single person.}
    \vspace{-0.2in}
    \label{fig:MyStyle}
\end{figure}

The MyStyle architecture \cite{melnik2022mystyle}\cite{nitzan2022mystyle} extends the idea of \textit{pivotal} fine-tuning from a single to hundreds of \textit{pivots}, all taken to be portrait images (called \textit{anchors}) of a given person. The authors propose to collect about 100 \textit{anchor} examples
that form a \textit{personalized region} in the latent space $\mathcal{W}$ of the StyleGAN2 generator and fine-tune the StyleGAN2 generator on these examples.
This allows it to use the StyleGAN2 generator as backbone for super-resolution (see Chapter \ref{ch:myStyle_Inpainting_SResolution}), inpainting or semantic editing tasks for a given person. 
This \textit{anchor} space was described by Generalized Barycentric coordinates, which allows us to detect when a latent point is inside the convex region of known \textit{anchors} in the latent space, and when a latent point is outside this region. 
For semantic editing task is necessary to preserve personality. Existing directions in the latent space are learned for the whole domain of faces and are not personalized. Because of this, some directions may lie outside the \textit{anchor} space and any small step in this direction will lead to degradation of the quality of the edited image. To solve this problem we need to project the direction on the \textit{anchor} space and perform editing in it \cite{nitzan2022mystyle}.

\subsubsection{HyperStyle: StyleGAN Inversion with HyperNetworks for Real Image Editing}
\label{inversion:hyperstyle}

\begin{figure}[!t]
    \centering
    \includegraphics[width=3.5in]{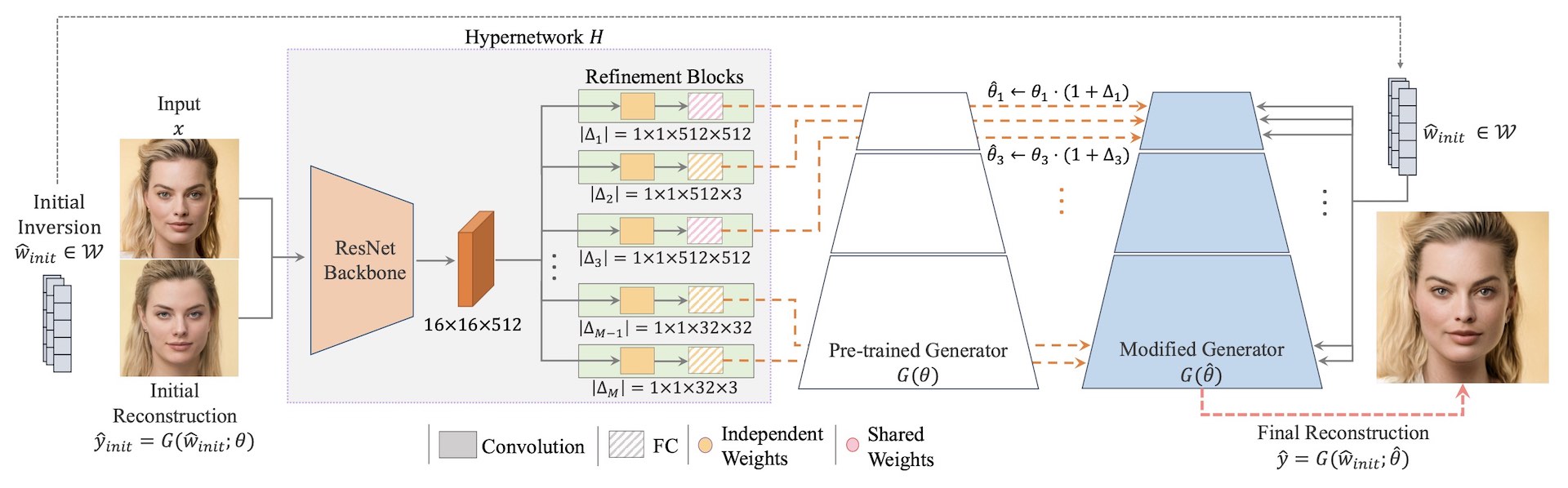}
    \vspace{-0.2in}
    \caption{HyperStyle method illustration \cite{alaluf2022hyperstyleNA}}.
    \vspace{-0.2in}
    \label{fig:Hyperstyle}
\end{figure}

The \textit{HyperStyle} \cite{alaluf2022hyperstyleNA} approach proposes to combine several ideas of the previous discussed approaches of \textit{ReStyle} \cite{alaluf2021restyle} (iterative improvement with inversion encoder), \textit{Pivotal Tuning} \cite{roich2022pivotal} (fine-tuning of StyleGAN generator weights for a specific image of a face), and \textit{e4e} \cite{Tov2021DesigningAE} (initial inversion prediction of $w \in \mathcal{W}$) for fine-tuning the pre-trained StyleGAN generator on the fly.

Key idea is to learn to predict coefficients for a channel-wise scaling of weights of selected layers of the pre-trained StyleGAN generator (see Figure \ref{fig:Hyperstyle}) to personalize it for a given image of a face. In this way \textit{HyperStyle} is computationally much lighter direct mapping for fine-tuning coefficients of the StyleGAN generator, than the gradient based fine-tuning approaches like \textit{Pivotal Tuning} \cite{roich2022pivotal}. The required scaling of channel weights is parameterized as
$
    \hat{\theta}_{l}^{i, j} := \theta_{l}^{i, j} (1 + \Delta_{l}^{i, j})
$, 
where $\theta_{l}^{i, j}$ denotes the weights of the $j$-th channel in the $i$-th filter in the StyleGAN generator's $l$-th layer and $\Delta_{l}^{i, j}$ is the output of the Refinement Blocks (see Figure \ref{fig:Hyperstyle}). For a better trade-off between network expressiveness and feasibility of learning, the \textit{HyperStyle} model is trained to predict scaling coefficients
only for kernels of selected StyleGAN generator layers. As the initial inversion captures coarse details, only layers that are responsible for medium and fine details may be selected.
The \textit{HyperStyle} architecture with iterative refinement of StyleGAN weights (depicted in  Figure \ref{fig:Hyperstyle}) consists of a \textit{ResNet} backbone that receives a source image with its initial reconstruction and Refinement Blocks of convolutions and fully-connected layer. Training is guided by an image-space reconstruction objective through pixel-wise $L_2$, LPIPS \cite{Zhang_2018_CVPR} and ArcFace \cite{deng2019arcface} losses.

\section{Editing Face Images with StyleGAN} \label{sec:editStyleGAN}

The main idea about editing an image using StyleGAN is that editing is achieved through some manipulation of its latent code, thereby moving the point that represents this latent code within one of the latent spaces of StyleGAN (see Chapter $\ref{chapter:latentspace}$). At first we do not know how movement in the latent space will affect the generated image but there are methods to learn how to navigate the latent Space to edit an image in a more controlled and semantically meaningful manner. Movement of the point in a wrong or a random direction will in the worst case lead away from the face distribution (thereby destroying the ''faceness'' of the image), or lead to an undesired, simultaneous change of different attributes, most likely accompanied by a loss of identity of a person on the image.

In this chapter we will focus on methods to identify directions in a latent space of StyleGAN which are correlated with desired, \textit{semantically interpretable editing attributes}, like smile, pose, hairstyle, age, and so on. Also, we will discuss how the different latent spaces of the StyleGAN architecture differ with regard to their editing properties.

Identifying semantic directions can typically be divided into global semantic directions (see Chapter \ref{ch:global_semantic_directions}) and single image semantic directions (see Chapter \ref{ch:single_image}).
Chapter \ref{ch:InterfaceGAN} introduces a supervised approach for finding semantic directions.
Chapter \ref{ch:GANspace} offers an unsupervised approach for discovering semantic directions by using PCA. Chapter \ref{ch:StyleSpace} describes discovering directions in StyleSpace using only few points with known attributes. Chapter \ref{ch:StyleCLIP} introduces a text-driven approach for discovering semantic directions.

\subsection{Global semantic directions} \label{ch:global_semantic_directions}

The first class of approaches is based on averaging a direction in latent space correlated with the given attribute over a number of contrastive pairs, with and without a desired attribute.

\subsubsection{Identifying Semantic Directions in a $\mathcal{W}$ space} \label{ch:InterfaceGAN}

InterFaceGAN \cite{shen2020interpreting}
proposes a supervised method for identifying semantic directions in a StyleGAN's latent space.
Suppose there is a pre-trained binary classifier for some attribute we want to edit. We make an assumption that for any binary semantic attribute there exists a hyperplane in the StyleGAN latent space serving as the separation boundary between points of positive and negative examples of that attribute. To find a suitable separating hyperplane that represents editing direction in the latent space $\mathcal{W}$ of the StyleGAN, InterFaceGAN \cite{shen2020interpreting} proposes to apply a linear SVM to points of positive and negative examples of the selected semantic attribute. 
They generate images for 500,000 randomly selected latent points in $\mathcal{W}$ and use the scores of a pre-trained ResNet binary classifier to identify a subset of 10,000 images for which the classifier reports the highest scores (attribute is present), and 10,000 images with the lowest scores (attribute is absent). Thus, to manipulate the attribute of an image, we can move the original latent point along the normal vector of the hyperplane.

When we edit several fine-grained attributes, one may affect another. To achieve a more disentangled editing it was proposed to orthogonalize a discovered set of semantic directions \cite{shen2020interpreting}. For example, given two hyperplanes for two semantics with normal vectors $n_1, n_2$, then we can find a projected direction $n = n_1 - (n_1^Tn_2)n_2$ such that editing along this projected direction correlates with the first attribute, but is not affecting the second attribute.

\subsubsection{Discovering Interpretable Semantic Directions in $\mathcal{W}$} %

\label{ch:GANspace}

\begin{figure}[!t]
    \centering
    \includegraphics[width=2.0in]{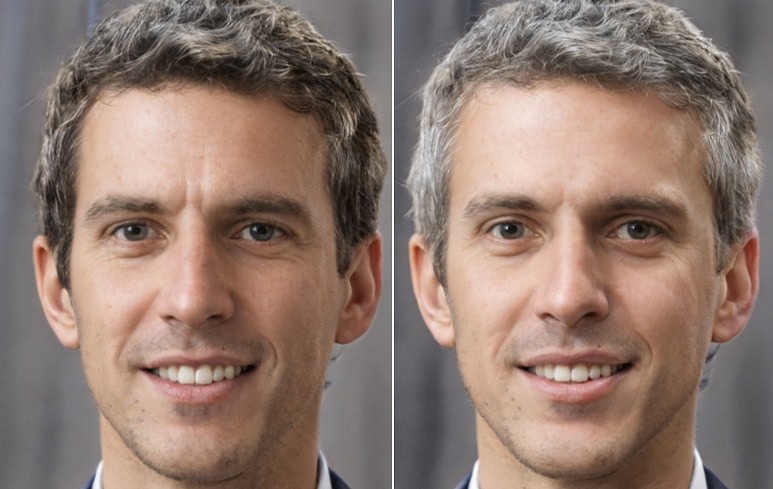}
    \vspace{-0.05in}
    \caption{Moving along the 10th principal component in 7-8 layers changes hair color \cite{harkonen2020ganspace}.}
    \vspace{-0.2in}
    \label{fig:pca}
\end{figure}

In contrast to the previous, supervised method, GANSpace \cite{harkonen2020ganspace} describes an unsupervised approach for discovering semantic directions. The method, which requires only a pre-trained StyleGAN generator, proposes sampling of a large number of random points in $\mathcal{Z}$ space, collecting the corresponding points in $\mathcal{W}$ space, and applying PCA to obtain a basis in $\mathcal{W}$ space. These PCA basis vectors contain the semantic directions responsible for some attributes. Within their study, using on the order of 100 basis vectors, the authors find that large-scale changes to geometric configuration seem to be limited to the first 20 principal components. For a further refined control of editing features, these PCA directions can be applied individually to each of the 18 StyleGAN2 layers. By checking the influence of the discovered PCA directions, it is possible to identify directions responsible for specific face editing attributes, for example hair color (see Figure \ref{fig:pca}). The authors show that such a check -- in their work of about 1800 combinations -- can be done both automatically or manually.

\subsubsection{Identifying Semantic Directions in StyleSpace ($\mathcal{S}$)}
\label{ch:StyleSpace}

The search for semantic directions can be done not only in the $\mathcal{W}$ space, but also in the more disentangled StyleSpace ($\mathcal{S}$ space, see Chapter \ref{ch:stylespace}) \cite{Wu2021StyleSpaceAD}. In this space, identification of semantic directions can already succeed with as few as 10 to 30 positive examples that contain the target attribute. This utilizes the idea that the differences between the mean style vector of the positive examples and the mean of the entire generated distribution of 500k samples in $\mathcal{S}$ space reveal which StyleGAN channels are the most relevant for the target attribute \cite{Wu2021StyleSpaceAD}.
Such channels can be identified, e.g., with statistical methods that determine which vector components show statistically significant deviations towards higher values.
Once we know the right channel, we can change its activation along the corresponding $\mathcal{S}$ dimension so that the generated image shows an increase or decrease of the desired semantic attribute.

\subsubsection{Text-Driven Discovering of Semantic Directions} %
\label{ch:StyleCLIPglobal}
\ndrww{
It is also possible to search for a global semantic direction using combined image-text embeddings, e.g., using the CLIP model \cite{radford2021learning}\cite{patashnik2021styleclip}.
For example, to find the semantic direction of the attribute \textit{smile} in the latent space of CLIP, we can take the direction between text embedding vectors such as \textit{face with smile} and \textit{face}.
To find the global direction of the example attribute \textit{smile} in the StyleSpace $\mathcal{S}$ of StyleGAN for some image, the coordinate of its latent point is moved in the positive and negative directions for a selected channel.
This produces a pair of generated images ($\pm \sigma$) that are fed into CLIP to measure how the perturbed direction of the chosen StyleGAN channel in $\mathcal{S}$ space is correlated with semantic direction of \textit{smile} in the latent space of CLIP. 
This process is repeated for a number of images always using the same channel in $\mathcal{S}$ space, to compute the corresponding averaged direction in CLIP space. Projection of this computed average direction on the semantic direction of attribute \textit{smile} in the latent space of CLIP provides a direct measure how much the chosen channel in $\mathcal{S}$ space affects the selected semantic direction in CLIP space.
After going through all the StyleGAN2 channels, those with projection values greater than a certain threshold are selected as related to the given semantic direction in the $\mathcal{S}$ space.
}

\subsection{Single image semantic directions} %
\label{ch:single_image}

The second class of approaches optimizes a single image and requires an assessment network over the generated image, like CLIP or a binary attribute classifier. Loss from such an assessment network produces a gradient step in latent spaces which indicates the direction to move the input point of the generator to decrease the target attribute loss.

\subsubsection{Text-Driven Manipulation of StyleGAN Imagery} %
\label{ch:StyleCLIP}

The \textit{StyleCLIP} \cite{patashnik2021styleclip} offers a text-based approach to image editing (technical details are already described in \ref{ch:StyleCLIPglobal}). 
The CLIP-model produces gradients through the CLIP image-encoder to minimize the similarity loss of text and image points in the combined latent space of CLIP, thus allowing to obtain the overall gradient that indicates the editing direction for the point in the $\mathcal{W}$ latent space of StyleGAN with regard to the textual attribute that is represented in CLIP. In addition, Identity \cite{deng2019arcface} and $L_2$ losses on the StyleGAN generated image allow to penalize loosing the person's identity in the generated image under text-attribute driven changes. However, the disadvantage of this method is the need to conduct computationally expensive gradient based optimization steps for each image and text attribute, because each pairing of image and text attribute gets its own editing direction in a StyleGAN latent space.

\begin{figure}[!t]
    \centering
    \includegraphics[width=3.5in]{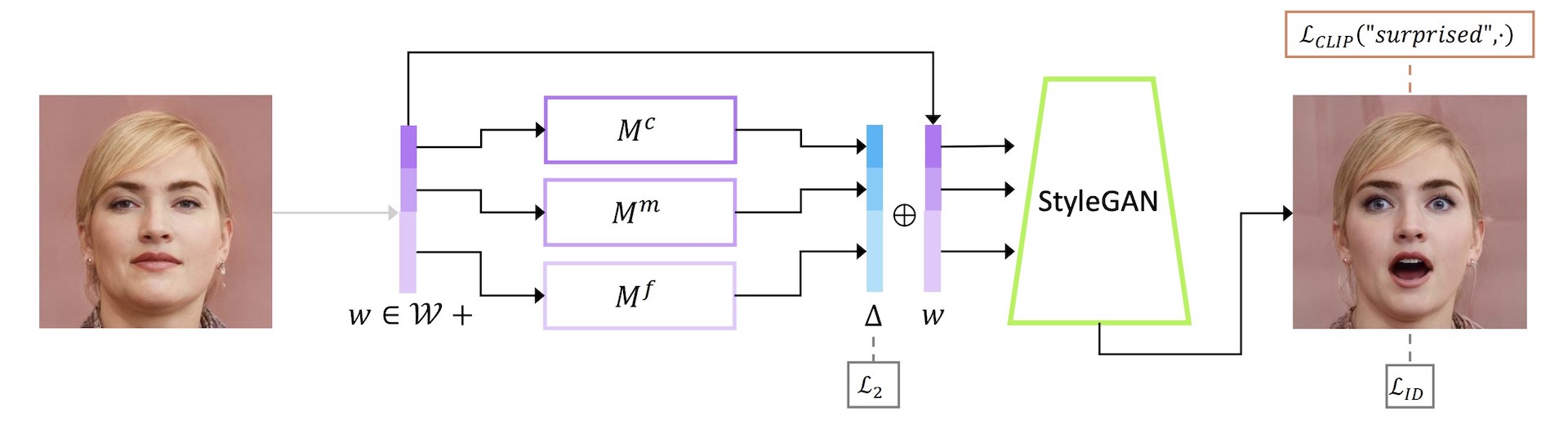}
    \vspace{-0.3in}
    \caption{\textit{StyleCLIP} mapping network architecture \cite{patashnik2021styleclip}. For a fixed "surprised" text attribute three mapping networks $M^c$, $M^m$ and $M^f$ for coarse, medium and fine scale are trained for images to map their latent vector $w$ (blue bar, left) of images into a short change direction $\Delta$ that, if added to $w$ (blue bars, right side), moves the generated output towards better satisfying (lower CLIP loss of the new image-text pair) the chosen text attribute.}
    \vspace{-0.1in}
    \label{fig:StyleClipMapping}
\end{figure}

As seen several times before, this problem of computationally expensive gradient based optimization can again be remedied by training an additional mapping network (Figure \ref{fig:StyleClipMapping}). Each text attribute that is of interest for editing, e.g. \textit{smile}, requires to train a separate mapping network. This \textit{StyleCLIP} mapping network \cite{patashnik2021styleclip} moves the point in $\mathcal{W+}$ space according to the text attribute. The \textit{StyleCLIP} mapping network employs an architecture with three prediction sub-networks (denoted as $M^c$, $M^m$ and $M^f$ in Figure \ref{fig:StyleClipMapping}) for the coarse, medium and fine layers of the StyleGAN2 generator respectively. Each sub-network takes the corresponding $\mathcal{W+}$ dimensions as input and predicts $\Delta_w$ for them (see Figure \ref{fig:StyleClipMapping}). The overall gradient guidance by the CLIP loss of the text-image pair is supplemented by terms to keep the result image in the vicinity of the input image (small $L_2$ loss) and to preserve identity (identity loss $L_{id}$). Authors employ the CelebA-HQ dataset \cite{Karras2018ProgressiveGONA} for training.

\subsubsection{Interpretable Control: Facial Pose and Expression Change Based on FaceRig}
\label{par:face_rig}
To be able to create editable faces as flawlessly as possible, controlling semantic face parameters that are interpretable in 3D like face pose and expressions play a crucial role. StyleRig  \cite{tewari2020stylerig} by Tewari et al. offers nice results using both three-dimensional morphable face models (3DMM) in combination with StyleGAN to provide such control.
(See Figure \ref{fig:image_grid_applications}K for StyleRig examples)

\section{Cross Domain Face Stylization} \label{ch:cross_domain}

The original StyleGAN architecture allows mixing styles of two sources of information (see Figure \ref{fig:style_mixing}) from the same training domain, for example CelebA-HQ dataset \cite{Karras2018ProgressiveGONA}. However, what if we want to mix styles of two sources of information from different domains, for example mixing with cartoon faces (Fig. \ref{fig:blendgan})? This chapter discusses approaches for style mixing from different domains by fine-tuning or merging the StyleGAN generators.

\denis{
StyleAlign \cite{wu2022stylealign} explains why we can do cross domain face stylization using StyleGAN fine-tuning to another domain (see Chapter \ref{domain:fine-tune}). The layer-swapping approach \cite{pinkney2020resolution} devises a controllable domain adaptation between original and fine-tuned StyleGAN models (see Chapter \ref{domain:layer-swap}). BlendGAN \cite{liu2021blendgan} proposes a style encoder for fine-tuning face StyleGAN such that its output becomes adaptable to an arbitrary style (see Chapter \ref{domain:blendgan}). StyleGAN-NADA \cite{DBLP:journals/corr/abs-2108-00946} proposes fine-tuning of the StyleGAN generator towards a target style domain using CLIP (see Chapter \ref{domain:stylgan-nada}). 
}

\denis{

\subsection{Fine-tuning generator to the target style domain}
\label{domain:fine-tune}
Let's consider two similar style domains - source and target. We can fine-tune the StyleGAN generator pre-trained on the source domain towards the target domain. To stylize a source image we can invert it into a latent code and generate from this code using the fine-tuned generator. StyleAlign \cite{wu2022stylealign} work suggests that the same latent code $z \in \mathcal{Z}$ is mapped to similar codes in $\mathcal{W}$ in source and target domains, which is called \textit{point-wise alignment}. Moreover, changing individual channels in $\mathcal{S}$ or moving in directions in $\mathcal{W}$ leads to the same semantic changes in generated images of source and target domains. This alignment can be measured by calculating the overlap between channels in $\mathcal{S}$ used for semantic editing (see Chapter \ref{ch:StyleSpace}) in original and fine-tuned generator, which is called \textit{semantic alignment} \cite{wu2022stylealign}. These alignments explain why fine-tuning based methods can be used for cross domain face stylization.

}

\begin{figure}[!t]
    \centering
    \includegraphics[width=3.5in]{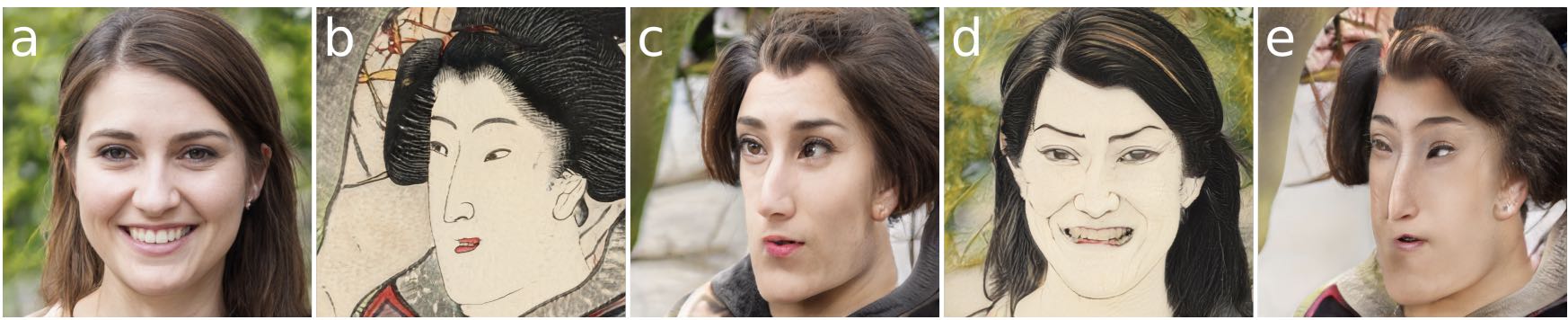}
    \vspace{-0.2in}
    \caption{\cite{pinkney2020resolution} (a) and (b) are samples of the ''Base'' and ''Transferred'' trained models, respectively. (c)(d)(e) are ''Interpolated'' results of different combination of layers from the ''Base'' and ''Transferred'' models.}
    \label{fig:layer-swapping-example}
\end{figure}

\denis{

\subsection{Layer swapping}
\label{domain:layer-swap}
Since we know that fine-tuning can be applied to cross domain stylization, we need methods to better control the stylization. \textit{Point-wise alignment} doesn't guarantee that we will preserve all the necessary features from the original face after sampling from a tuned generator. So in \cite{pinkney2020resolution} a layer swapping technique was proposed (see Figure \ref{fig:layer-swapping-example}). This technique allows to control which features we want to transfer from a target domain.

Layer swapping domain adaptation is done by combining selected layers from the original and fine-tuned generator. Now we can transfer, for example, low level features from the target domain such as face geometry or hair style (see Figure \ref{fig:layer-swapping-example}C) or transfer only high level features (see Figure \ref{fig:layer-swapping-example}D).

If it's not enough to swap layers, for example, when the shape variations between the photo domain and the target domain are too large to successfully capture the exaggerations and color stylization of caricatures using the layer swapping technique, we can use additional blocks\cite{jang2021stylecarigan} appended to the layers of the first StyleGAN blocks to modify coarse features.

}

\subsection{Blending for Stylized Face Generation} %

\label{domain:blendgan}

\begin{figure}[!t]
    \centering
    \includegraphics[width=3.5in]{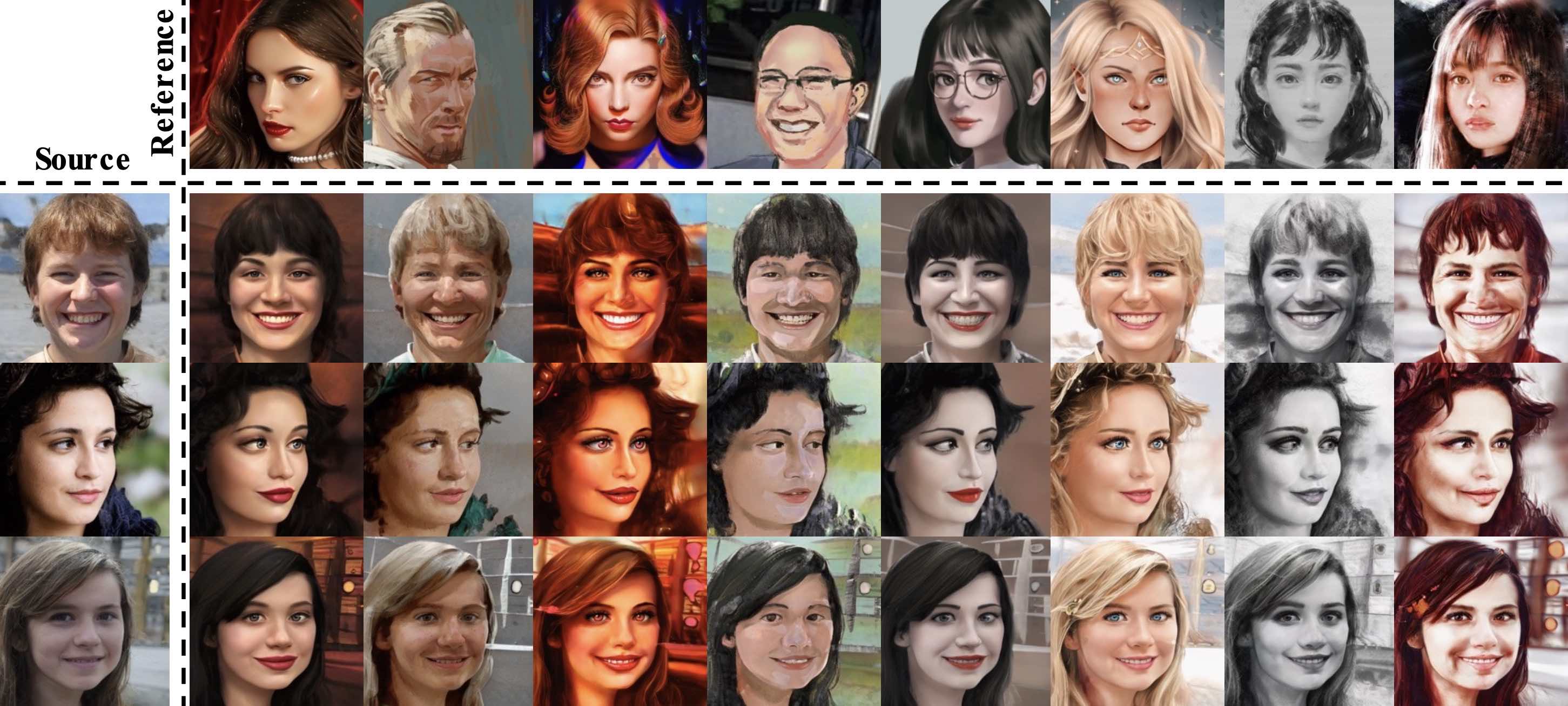}
    \vspace{-0.2in}
    \caption{Examples of reference-guided BlendGAN synthesis \cite{liu2021blendgan}.}
    \vspace{-0.2in}
    \label{fig:blendgan}
\end{figure}

\denis{

Paper \cite{liu2021blendgan} suggested an alternative to the layer swapping mechanism which allows to generate images in different domains using only one model. It proposed to fine-tune a StyleGAN model on a dataset of faces in different artistic styles (see Figure \ref{fig:blendgan}). A separate encoder is trained using contrastive learning to extract style latent code $z$ from artistic images. The encoder is trained on augmented variations of stylized images of faces to produce the same latent vector for geometric augmentations of the same style image, but distinct to latent vectors of other style images. Thus, the encoder becomes insensitive for facial details, but focused on the style of an artistic face image. The latent space of this encoder is connected to the latent space $\mathcal{W}$ of StyleGAN via a second mapping network of 8 fully connected layers. 

The blending procedure is similar to the vanilla StyleGAN merging of two sources of information. The resulting image is generated by taking two points in the latent space $\mathcal{Z}$. One $z$ point that represents a face is being transforming into the latent space $\mathcal{W}$ using the original mapping network and used for some first layers of the StyleGAN model. The second $z$ point that represents a style is obtained from the reference artistic image by the encoder. This point is being transformed into the latent space $\mathcal{W}$ using the second mapping network and used for other layers of StyleGAN.

}
\subsection{CLIP-Guided Domain Adaptation of StyleGAN}
\label{domain:stylgan-nada}

\begin{figure}[!t]
    \centering
    \includegraphics[width=2.0in]{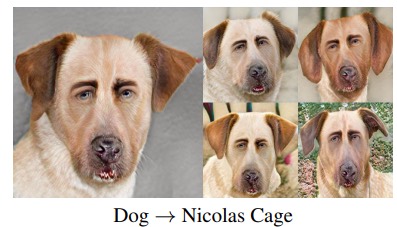}
    \vspace{-0.1in}
    \caption{StyleGAN-NADA \cite{DBLP:journals/corr/abs-2108-00946} allows to create images that do not even exist in real life ("Nicolas Cage Dogs" using a generator trained to generate dogs and the "Nicolas Cage" text prompt).}
    \label{fig:NicolasCageDog}
\end{figure}

\denis{Default fine-tuning and layer swapping requires a large dataset of stylized images from the same domain. There are some advanced methods for fine-tuning a generator to a target domain without large stylized datasets}. StyleGAN-NADA \cite{DBLP:journals/corr/abs-2108-00946} allows to fine-tune the StyleGAN generator using CLIP based gradients towards a different style domain using only text prompts, for example "zombie faces". For that, first the semantic direction of the prompt is identified in CLIP latent space by subtracting embeddings of two text prompts "zombie face" and "face". Using
the CelebA-HQ dataset \cite{Karras2018ProgressiveGONA} for training, the authors demonstrate that the
semantic direction in CLIP latent space can be used to fine-tune a StyleGAN model \cite{Karras2018ProgressiveGONA} towards domain defined by the text. To this end, thousands of images of faces are generated by sampling a latent space of the trained StyleGAN model. Then, according to the loss, embeddings of these images in CLIP space have to move in a parallel direction to the found semantic direction in CLIP text space. This parallel movement of thousands of points in the CLIP latent space becomes the training objective for changes of the StyleGAN model weights.

\dennis{

}

To solve overfitting and divergence problems, \cite{DBLP:journals/corr/abs-2108-00946} proposes to fine-tune only StyleGAN layers with the most influence on the new domain and freeze weights of the rest of the layers. To determine fine-tuned layers, the approach moves these latent image points in CLIP space to the text embedding of the target domain and chooses those layers in which the latent code has changed the most. These layers are considered as sensitive to the new domain. Further only these layers are unfrozen and generator weights are fine-tuned by above mentioned method.

\dennis{

}

\section{Face Restoration with StyleGAN} \label{ch:FaceRestoration}

Face restoration aims at recovering high-quality (HQ) face detail from low-quality (LQ) counterparts suffering from unknown degradation, such as low resolution, blur, compression artifacts or noise. Context information may allow to properly infer such missing detail. For example, location of hairs on a blurry face can be estimated from low resolution details of a face. Exact location of each hair is not important for human perception, but it is important to restore overall structure and texture. Thus one may expect that supervised training of neural network architectures with LQ–HQ training pairs using pixel-wise losses will not lead to high perceptual quality. Instead, pixel-wise loss functions cause over-smoothed result images, as the model tends to be the mean of high-quality faces.

However, architectures with energy-based components, like GAN discriminators, can be a good approach for reconstruction of fine detail from low resolution images and may provide perceptually plausible results.
In the fully converged state, the discriminator will not be able to differentiate between a generated sample and a sample from the dataset. Ideally, the same will then also apply to human perception. Thus, we can expect GANs to be able to generate deblurred images of perceptually high-quality, but there remains the challenge of mapping a blurred input into an appropriate latent vector of StyleGAN that makes StyleGAN generate the corresponding deblurred version.

\subsection{GAN Prior Embedded Network for Blind Face Restoration in the Wild}
\label{rest:gpen}

\begin{figure}[!t]
    \centering
    \includegraphics[width=3.5in]{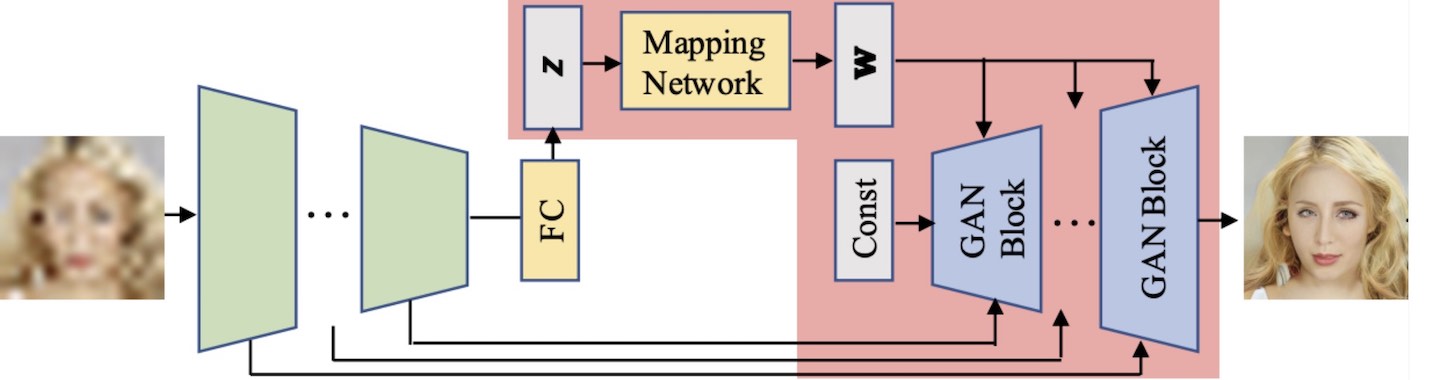}
    \vspace{-0.2in}
    \caption{GPEN architecture \cite{yang2021gan}. A CNN encoder network receives a LQ input (left) and has learned to provide $z$-input and via skip connections control inputs to additional input fields in the GAN generator layer hierarchy (right) to cause generation of a corresponding HQ output image.}
    \label{fig:gpen}
\end{figure}

The GAN Prior Embedded Network (GPEN)\cite{yang2021gan} is one of the solutions for blind face restoration problem. The main idea of this approach is to embed a StyleGAN like architecture as the decoder part into an encoder-decoder architecture, thereby utilizing its strong prior for generating high-quality faces. The LQ image is fed to the encoder part, which is realized as a conventional CNN whose output serves as the latent code for the subsequent
GAN decoder to generate an appropriate HQ image (see Figure \ref{fig:gpen}). To this end, the activations of the final CNN output layer are fed in place of the random vector into the GAN's latent space mapping network. Furthermore,
to make efficient use of the controllability
of the StyleGAN decoder, the GAN is slightly modified by padding its generator layers with additional input fields for receiving activity patterns via skip connections from more shallow layers in the encoder (see Figure \ref{fig:gpen}). They are chosen such that the feature map hierarchy of the encoder CNN and the GAN generator layers become connected at matching levels of resolution.

In the first phase the GAN network (for example the StyleGAN generator) is trained to generate high quality faces from scratch. This phase uses the FFHQ dataset. During this phase, the added channels of the GAN layers are not yet used by the encoder and the GAN generator network is encouraged to learn ignoring signals from these noise channels by just providing random noise for these inputs. Thus, in this phase the noise channels play a placeholder role for a subsequent fine-tuning of the GAN prior network. In the second phase of training, the encoder is included, now replacing the random inputs to the GAN by the activations of matching encoder layers.

For the second phase, training the whole architecture, a dataset of  pairs of low quality (LQ) and high quality (HQ) images was synthesized. HQ images are downgraded to obtain LQ images using blurring, gaussian noise, downsampling and JPEG compression. A weighting of three losses was used to train the GPEN architecture: $L_1$, adversarial and LPIPS losses (see Chapter \ref{sim:adversarial} and \ref{sim:lpips}) based on features from discriminator, rather than VGG network.

\subsection{GFP-GAN: Towards Real-World Blind Face Restoration with Generative Facial Prior}
\label{rest:gfpgan}

\begin{figure}[!t]
    \centering
    \includegraphics[width=3.5in]{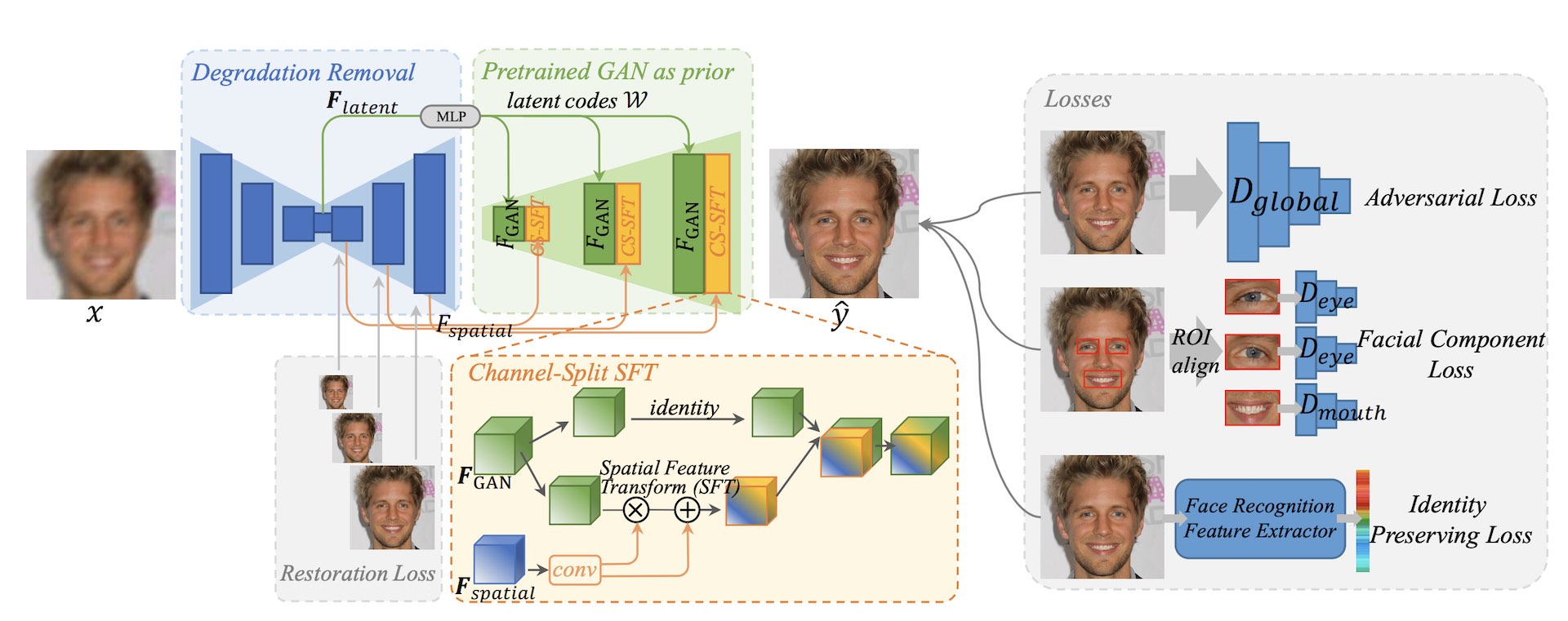}
    \vspace{-0.2in}
    \caption{GFP-GAN architecture \cite{wang2021towards}}
    \vspace{-0.15in}
    \label{fig:gfpgan-arch}
\end{figure}

Generative Facial Prior GAN (GFP-GAN) \cite{wang2021towards} is a framework for real-world blind face restoration applications (see Figure \ref{fig:gfpgan-arch}). It resembles GPEN (see Chapter \ref{rest:gpen}), but employs a degradation removal module in the form of a U-Net as an encoder, along with with a number of connectivity changes
that allow the usage of a pre-trained StyleGAN as a decoder without the need for a training from scratch and subsequent fine-tuning of a modified StyleGAN, as in GPEN.

To achieve this, the U-Net decoder features are fed to a channel split-spatial feature transform (CS-SFT) module that learns to modify activations in StyleGAN layers for better perceptual quality of the generated image with identity preservation. The interaction between the standard StyleGAN and the newly introduced spatial channels are shown in the CS-SFT box within Figure \ref{fig:gfpgan-arch}.
The SFT operation happens as follows. Spatial features from the U-Net go into a convolutional module generating an output tuple ($\alpha$, $\beta$) with parameters for spatial-wise feature modulation. Parameter $\alpha$ is being used for element-wise scaling of GAN features, $\beta$ specifies an additive shift. After this, the obtained modulated channels
are concatenated with the untouched GAN channels to form the final result, as depicted in box ``Channel-Split SFT'' in Figure \ref{fig:gfpgan-arch}.

Again a composition of losses is applied to train the entire architecture. First, motivated from the U-Net, a pixel-wise reconstruction loss $L_1$ between the degraded input image and ground truth version of the image at different scales (see the blue U-Net in Figure \ref{fig:gfpgan-arch}) is used. Second, the restored output image $\hat{x}$ of the StyleGAN generator can be compared with the degraded input image $x$ using LPIPS, and Identity losses (see Chapter \ref{ch:sim}). Third, since the eyes and mouth are crucial to a person's personality, a further Facial Component loss is introduced which consists of local discriminators for the left eye, right eye, mouth, and the complete face. The Region of Interest Align operator provides these discriminators with corresponding regions of the generated face.

\subsection{MyStyle: A Personalized Generative Prior}
\label{ch:myStyle_Inpainting_SResolution}

\begin{figure}[!t]
    \centering
    \includegraphics[width=3.5in]{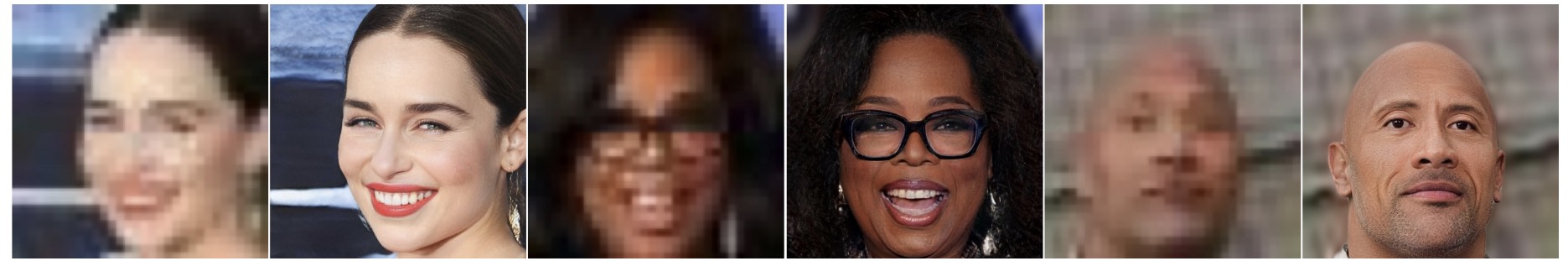}
    \vspace{-0.2in}
    \caption{Personalized super-resolution results using MyStyle \cite{melnik2022mystyle}\cite{nitzan2022mystyle}}
    \vspace{-0.15in}
\end{figure}

We have described this architecture \cite{melnik2022mystyle}\cite{nitzan2022mystyle} already in some detail in Chapter \ref{inversion:myStyle}. Here, from the perspective of HQ face image restoration, we only remark that this architecture also applies a StyleGAN to solve the problem of super-resolution, using $L_2$ pixel and LPIPS losses, with a downscaling operation applied to the images before application of these losses.

\subsection{VAEs with Codebooks - Alternative Approaches}
\denis{
Codeformer\cite{zhou2022codeformer} and VQFR\cite{gu2022vqfr} are encoder-decoder based approaches that don't use StyleGAN infrastructure, however achieve similar blind face restoration results. 
One common property of these approaches is that they use a discrete latent space (VQ-VAE codebooks \cite{DBLP:journals/corr/abs-1711-00937}). Also the spatial resolution of the bottleneck part and modulated skip connections have an important role in restoration-identity preservation trade-off.

}

\section{Deepfakes} \label{ch:deep_fake_chapter}

\begin{figure}[!t]
    \centering
    \includegraphics[width=3.5in]{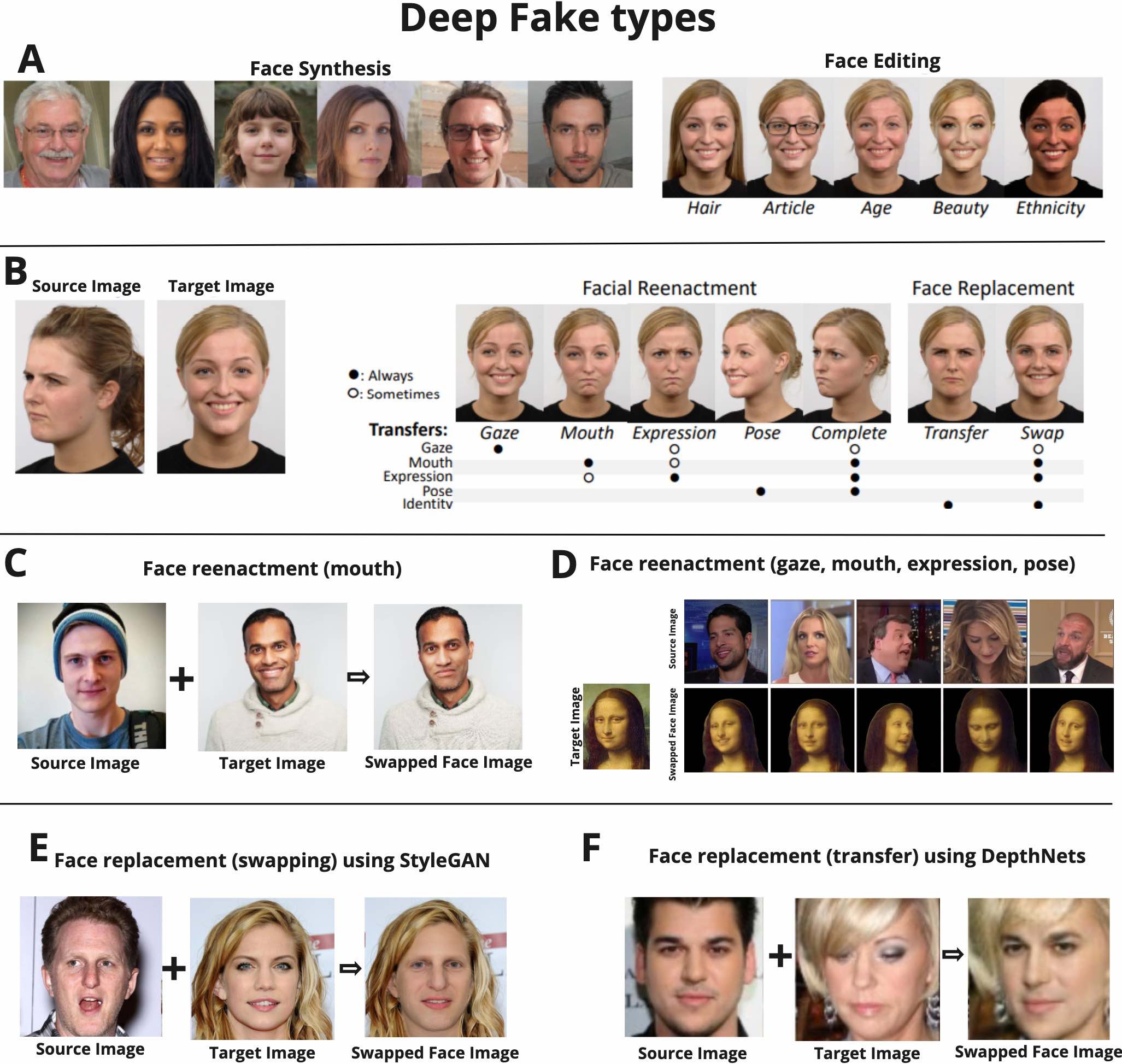}
    \vspace{-0.2in}
    \caption{A) Face Synthesis and editing examples \cite{mirsky2021creation}, B) Description of the types of deepfakes \cite{mirsky2021creation}, C) Face Reenactment of the mouth \cite{Yang2019UnconstrainedFE}, D) complete face reenactment \cite{9157131}, E) Face swapping utilizing StyleGAN \cite{Zhu2021OneSF}, F) Face Transfer using DepthNets \cite{moniz2018unsupervised}.}
    \label{fig:deep_fakes}
\end{figure}

\textit{Deepfake} is the manipulation of an image of a face that is hardly recognizeable by humans. Deepfake creation methods can be classified into the following types (Fig. \ref{fig:deep_fakes}):

\textbf{Synthesis:} a deepfake face is created without a face image. Chapter \ref{sec:NNarch4FaceGen} covers the synthesis of photo-realistic fake images of faces that do not exist.

\textbf{Editing:} facial features are altered, added or removed (see Chapter \ref{sec:editStyleGAN}).

\textbf{Reenactment:} a source face is used to control gaze, mouth or expression of a target face \cite{thies2016face2face}\cite{Tulyakov2018MoCoGANDM}\cite{9157131}. Following articles focus on mouth reenactment \cite{song2019talking}\cite{8970886}, gaze reenactment \cite{8010348}, and pose reenactment \cite{Tran2019RepresentationLB}.

\textbf{Replacement:} the identity of the source face is used to replace the identity of the target face \cite{fake_app}\cite{Zhang2019FaceSwapNetLG}\cite{9156865}.

\textbf{Replacement (Transfer):} the identity and expression of the source face is used to replace the identity and expression of the target face, which can be perceived as consecutive face swapping and face reenactment \cite{moniz2018unsupervised}.

This chapter will focus on \textbf{reenactment} (Chapters \ref{ch:Facial_Reenactment}) and \textbf{replacement} (Chapter \ref{ch:Facial_Replacement}) methods. The main difference between face reenactment and face replacement is that in face replacement, the identity of the source face is transferred to the target face.

\subsection{Face Reenactment}
\label{ch:Facial_Reenactment}

Face reenactment (see Figure \ref{fig:deep_fakes}B) is useful in the advertisement or in the film industry \cite{mirsky2021creation}. Typically, if the main actor (target) has limited time or high costs, another substitute actor (source) can perform the acting and the expressions are then transferred to the main actors face, as if he or she is actually performing \cite{screenRant}. In this case, the main actor's face would be the target face and the substitute actor would be the source.
One usage of mouth reenactment could be voice dubbing into another language for advertisements, music or video games.
Gaze reenactment can be used to improve photographs by making the people (target) look into the camera (or where desired) \cite{Lip-syncing}.
Pose reenactment has been used for face frontalization in security footage (target) to improve face recognition \cite{LIU2022103526}.

Facial reenactment can be realized by StyleGAN and its latent code. For instance,
for transferring expression from a source image $x_s$ to a target image $x_t$ the authors of \cite{Yang2019UnconstrainedFE} solve the following optimization problem to find the latent code $s_{r}$ of
the desired result image:
\begin{equation*}
    s_{r}=\operatorname*{argmin}_s [Dist_1(G(s), x_s) + Dist_2(G(s), x_t)].
\end{equation*}

Here, $G(\cdot)$ is the mapping function of the generator,
and $Dist_1(x,x'),Dist_2(x,x')$ are suitable distance functions
in image space that measure similarity of a face image pair
w.r.t. expression ($Dist_1$) and appearance ($Dist_2$) of the faces.

To solve this equation the authors constrain the solution space to particular linear combinations $s_{r}=\alpha s_s + \beta s_t$ of the StyleGAN latent codes of the given images. Here, $\alpha$ and $\beta=\textbf{1}-\alpha$ are diagonal
$18 \times 18$ matrices whose diagonal elements may only attain values
0 or 1. This is equivalent to adopting for the 18 style parameters of $s_{r}$ either the corresponding
value of $s_s$ or $s_t$, leading to a finite search space
of $2^{18}$ possibilities for the optimization problem.
From their solutions, they discover that StyleGAN layer
4 is primarily responsible for expression, while other
layers (e.g., 8 and 9) primarily control hair color and
hat style \cite{Yang2019UnconstrainedFE}.

\subsection{Face Replacement and Face Transfer}
\label{ch:Facial_Replacement}

Face replacement and face transfer are closely related. While face replacement swaps a face into the position of a previous face, face transfer additionally takes care to make the swapped face inherit the expression of the previous face, which is usually achieved with an additional face reenactment step.

Face replacement can be established using classical computer graphics-based approaches \cite{8124497} by using facial landmarks extracted for each face. The results, however, are not as good as using neural network based methods.

FakeApp \cite{fake_app} used an encoder-decoder NN architecture to swap faces of two different people A and B. Expressions and emotions from the target image are kept, while the facial identity is swapped. First, it collects a dataset of faces of two people, A and B, using an object detection method \cite{limberg2022yolo}. Secondly, it trains two auto-encoders $E_A$, $E_B$ to encode and two decoders $DC_A$, $DC_B$ to reconstruct the faces of A and B respectively. The key idea is that the two encoders have to share the same weights, but they keep their respective decoder weights independent. This allows the encoder to learn global features of the two faces A and B while the two decoders are trained to use this general encoding to generate face specific details of person A or B respectively. Thirdly, to create the face swap, trained encoder $E_A$ and decoder $DC_B$ are used for the images of person A. This method is applicable for videos as well. The results are similar to Figure \ref{fig:deep_fakes}E and F. The downside of this approach is the large dataset of faces of both people that is needed to train the encoder-decoder networks. When using this method on videos, a big problem is the temporal coherence. As this method does not take preceding frames into account, it may produce flickering of the faces. This can be mitigated by providing context or implementing temporal coherence losses \cite{Vougioukas2018EndtoEndSF}\cite{Wang2018VideotoVideoS}.

The Face Transfer Module (FTM) \cite{Zhu2021OneSF} mitigates most of the problems described above. It can transfer faces without ever having seen them before. The expression of the face of the target image remains, but the facial identity is transferred from the target image (see Figure \ref{fig:deep_fakes}D). It uses a trained StyleGAN model as a decoder and employs the Hierarchical Representation Face Encoder (HieRFE), \cite{Zhu2021OneSF} which is based on the ResNet50 network, to project images of faces into the improved latent space of StyleGAN with transformers \cite{Li2022TransformingTL}.

\revise{\textit{Styletalk} method \cite{ma2023styletalk} creates lifelike animated faces from a single image of a speaker based on reference videos using 3D Morphable Models expression parameters.} Another approach \cite{moniz2018unsupervised} tries to predict the 3D pose of the face and its facial landmarks from the 2D source image and predict the rotation and translation of the landmarks to map them onto the facial landmarks of the target image. For predicting the 3D poses of the facial landmarks, \cite{moniz2018unsupervised} uses an unsupervised Siamese-like network \cite{10.5555/2987189.2987282} consisting of convolutions, pooling layers, and densely connected layers that they named \textit{DepthNet}. The two images pass through the network which then predicts the depth and affine transformation to map one face onto the other. A CycleGAN \cite{DBLP:journals/corr/ZhuPIE17} is used to cleanly blend the new face position into the image.

\ndrw{
\section{Alternative Face Generation and Editing approaches}
\label{ch:Alternative}

\subsection{3D-Consistent Generative Adversarial Networks}

NeRF \cite{mildenhall2021nerf} based approaches exhibit consistency of an object while altering position and camera orientation. Development of successive StyleGAN architectures goes along the improving of such consistency along different camera poses. Thus, a straightforward idea is to substitute the StyleGAN generator with a NeRF-like architecture with disentangled camera pose representation and integrate the idea of the mapping network and $W$ latent space \cite{gu2021stylenerf}. While the high resolution of StyleGAN ($1024 \times 1024$) can be computationally expensive for training a NeRF architecture that would need to compute a batch of 1M rays gradient updates, \cite{gu2021stylenerf} proposed to generate an only ($64 \times 64$) image/feature space with NeRF and upscale it to $1024 \times 1024$ using a generator modulated by W. Another 
performance
improvement is proposed by \cite{or2022stylesdf} where the encoding of positional information is proposed to be represented by a Tri-plane Decoder, thus reducing NeRF to learn direction information for every position. To conclude incorporating of the NeRF components allows to achieve the required consistency of face generation for different poses and directions, while building on solutions accumulated through the development of the StyleGAN architectures.

\subsection{Diffusion Models}

\textit{Deepfake} diffusion models \cite{bigioi2023speech} \cite{shen2023difftalk} \cite{stypulkowski2023diffused} have emerged as a result of advancements in diffusion-based methods. They are designed to generate realistic \textit{Deepfake} videos with audio information used as a conditioning signal. The person's identity is preserved in the generated videos, conditioned by a single image of a face or a full video, and the movements of their lips and other facial features are synchronized with the audio. The U-Net-based model serves as the diffusion component.

\textit{Face editing} with diffusion models can also be conditioned by text instruction. In this way the \textit{InstructPix2Pix} model \cite{brooks2022instructpix2pix} can edit images, including faces. To train such text-conditioned editing model, a dataset of triplets containing the original image, the edited image, and the text editing instruction, was generated \cite{brooks2022instructpix2pix}.
Similarly, \textit{ControlNet} \cite{zhang2023adding} allows to control the generation of images (including faces) from two sides: on the one hand the generated image originates from a simplified representation of a given image (Canny Edge, Semantic Segmentation or Normal Map), and on the other hand it must fit a certain text prompt. The text prompt can be changed, thus allowing image editing. \revise{In papers \cite{huang2023collaborative} \cite{nair2023unite}, pre-trained diffusion models are used to perform multi-modal guided face generation or editing. 
}

}

\ndrw{

\section{Conclusion}\label{ch:Conclusion}

StyleGAN architectures \cite{karras2019style}\cite{karras2020analyzing}\cite{karras2021alias} have been extensively researched as deep learning models for face generation and editing, making them the most studied in this domain. %
They are providing a versatile editing control over the generation process of high-quality images, which makes StyleGAN architectures particularly valuable.

Existing limitations of StyleGAN, and emerging competing techniques, like diffusion models, result in the following trends and expectations for future research in the area of face generation and editing with deep neural networks:

\begin{itemize}

\item \textbf{Foundation Models}: Pre-trained StyleGAN models can be considered as \textit{Foundation Models} and require further research on fast fine-tuning techniques, like neural network adapters for integration with other architectures \cite{zhang2023llama} \cite{sevastopolsky2023boost} \cite{wu2023describe3d}, scaling StyleGAN to large diverse datasets \cite{Sauer2022StyleGANXLSS} \revise{or new interactive image manipulation methods \cite{pan2023drag}}.

\item \textbf{Mobile Applications}: Generative models on-mobile-device are an open research topic with a lot of challenges. A StyleGAN model is too computationally demanding for mobile devices or applications that require a high frame rate for videos. One common solution is to employ StyleGAN to produce a synthetic dataset for a specific single feature, like pairs of face with and without sunglasses. Then, this synthetic dataset is utilized to train a tiny, single-function-image-processing network for mobile applications (see Section \ref{sec:Mobile_Networks}). Another solution is to distill a downscaled version of StyleGAN architecture from the original StyleGAN \cite{jia2023blazestylegan}. For example, \textit{BlazeStyleGAN} \cite{jia2023blazestylegan} has around 16 times fewer parameters and around 16 times fewer FLOPs needed to generate a $1024\textsuperscript{2}$ image than the StyleGAN architecture. As a result, \textit{BlazeStyleGAN} is able to generate high quality images on mobile devices. However, further research is needed to find out how latent space works in compact models compared to the original model.

\item \textbf{Diffusion Models}: are emerging competing approaches for face generation \cite {melzi2023gandiffface} \cite{kansy2023controllable} \cite{kim2023dcface}, editing \cite{bigioi2023speech} \cite{shen2023difftalk} \cite{stypulkowski2023diffused}, enhancement \cite{wang2023exploiting} \cite{liang2023iterative} \cite{wang2023dr2} and temporal consistency \cite{kim2023diffusion}. However, the evaluation step of such models is even more computationally demanding. Additionally, the control over the image generation and editing is so far not as well optimized as those of StyleGAN architectures. \revise{In addressing the aforementioned challenges, this work\cite{avrahami2023blended} presents an accelerated solution specifically tailored for the task of local text-driven editing of images.}

\item \textbf{Text2Face}: \denisnew{
Recent works \cite{sauer2023stylegan} \cite{kang2023gigagan} have shown that StyleGAN-based networks can be comparable in quality to diffusion models in Text2Image task, while having better performance. However, the quality of the generated images is still inferior to such strong works as Imagen \cite{Saharia2022PhotorealisticTD} or DALLE-2 \cite{Ramesh2022HierarchicalTI}, so an open question is to improve the generation of images from text using GANs.
}

\item \textbf{Neural Rendering}: The 3D consistency of a generated face when viewed from different angles is another crucial aspect of face perception. NeRF \cite{mildenhall2021nerf} is a technique that has shown impressive results for 3D scenes with faces \cite{zheng2023neuface}, and there have been efforts to combine StyleGAN and NeRF \cite{gu2021stylenerf} \revise{or to create 3D-GANs \cite{chan2022efficient} \cite{abdal20233davatargan} \cite{sun2023next3d} \cite{ma2023otavatar} and solve inversion task for them\cite{xu2023n}  \cite{yin20233d}}. 

\item \textbf{Temporal Consistency}: A number of works in this survey demonstrate nice style transfer results. However, when applied to a video sequence of frames, inconsistency of identity of a person, lighting conditions, and features flickering at different frames become obvious. \denisnew{Thus, it is necessary to apply different flicker suppression losses \cite{yang2022vtoonify}. Also good direction of development is to work with unaligned faces as in StyleGANEX \cite{yang2023styleganex}, which allows you to edit all the details of a video frame (hair, clothes, environment) in a compatible manner.}

\end{itemize}

We expect to see a multitude of forthcoming papers in the coming years focusing on the topics covered in this survey, but with an emphasis on applying them to a blend of StyleGAN, Diffusion, and NeRF techniques.

}

\bibliographystyle{IEEEtran}
\bibliography{citations.bib}

\end{document}